\documentclass[twoside]{article}
\usepackage[accepted]{aistats2018}
\usepackage{graphicx}
\usepackage{amsmath, amsthm, amssymb, diagbox}
\usepackage{rotating}
\usepackage{stackengine}
\usepackage{caption}
\usepackage{subcaption}

\usepackage{algorithm,algorithmic}
\usepackage{tikz}
\usetikzlibrary{fit,automata,positioning,calc}

\usetikzlibrary{positioning}
\usetikzlibrary{shapes,fit}
\tikzset{main node/.style={circle,fill=blue!20,draw,minimum size=1cm,inner sep=0pt}}

\newtheorem{corollary}{Corollary}

\theoremstyle{plain}

\theoremstyle{plain}

\theoremstyle{plain}
\newtheorem{lem}{\protect\lemmaname}
\theoremstyle{plain}

\theoremstyle{plain}
  
\theoremstyle{definition}

\theoremstyle{definition}

\theoremstyle{definition}

\providecommand{\claimname}{Claim}
\providecommand{\lemmaname}{Lemma}
\providecommand{\propositionname}{Proposition}
\providecommand{\theoremname}{Theorem}
\providecommand{\corollaryname}{Corollary} 
\providecommand{\definitionname}{Definition}
\providecommand{\assumptionname}{Assumption}
\providecommand{\remarkname}{Remark}

\DeclareMathOperator*{\argmax}{arg\,max}

\newcommand{\GP}{\mathrm{GP}}

\newcommand{\Bernoulli}{\mathrm{Bernoulli}}

\newcommand{\xv}{\mathbf{x}}

\newcommand{\yv}{\mathbf{y}}

\newcommand{\Gc}{\mathcal{G}}

\newcommand{\Xc}{\mathcal{X}}

\usepackage[letterpaper, pass]{geometry}

\begin{document}

%

%
\runningauthor{Paul Rolland, Jonathan Scarlett, Ilija Bogunovic, and Volkan Cevher}

\runningtitle{High-Dimensional Bayesian Optimization via Additive Models with Overlapping Groups}

\twocolumn[

\aistatstitle{High-Dimensional Bayesian Optimization via \\ Additive Models with Overlapping Groups}

\aistatsauthor{
 
    Paul Rolland\textsuperscript{\textnormal{1}}, Jonathan Scarlett\textsuperscript{\textnormal{2}}, Ilija Bogunovic\textsuperscript{\textnormal{1}}, Volkan Cevher\textsuperscript{\textnormal{1}} \\
    \textsuperscript{1} Laboratory for Information and Inference Systems (LIONS), EPFL \\
    \textsuperscript{2} Department of Computer Science \& Department of Mathematics, National University of Singapore \medskip \\ 
    {\{paul.rolland, ilija.bogunovic, volkan.cevher\}@epfl.ch}, {scarlett@comp.nus.edu.sg}
    \bigskip
    
}
]

%

\begin{abstract}
Bayesian optimization (BO) is a popular technique for sequential black-box function optimization, with applications including parameter tuning, robotics, environmental monitoring, and more.  One of the most important challenges in BO is the development of algorithms that scale to high dimensions, which remains a key open problem despite recent progress.  In this paper, we consider the approach of Kandasamy {\em et al.}~(2015), in which the high-dimensional function decomposes as a sum of lower-dimensional functions on subsets of the underlying variables.  In particular, we significantly generalize this approach by lifting the assumption that the subsets are disjoint, and consider additive models with {\em arbitrary} overlap among the subsets.  By representing the dependencies via a graph, we deduce an efficient message passing algorithm for optimizing the acquisition function.  In addition, we provide an algorithm for learning the graph from samples based on Gibbs sampling.  We empirically demonstrate the effectiveness of our methods on both synthetic and real-world data.
\end{abstract}

\vspace*{-3ex}
\section{Introduction}
\vspace*{-1ex}
Bayesian optimization (BO) is a powerful method for sequentially optimizing an unknown function $f$ that is costly to evaluate, for which noisy point evaluations are available.  Since its introduction, BO has successfully been applied to a variety of applications, including algorithm parameter tuning (e.g., deep neural networks) ~\cite{NIPS2012_4522},~\cite{bergstra:hal-00642998},~\cite{Mahendran:2012} and robotics~\cite{07ijcai-gait},~\cite{Martinez-Cantin-RSS-07}.  However, most successful applications of BO have involved low-dimensional input spaces.  Efficiently scaling to high dimensions remains a key open challenge, and is crucial in applications such as computer vision~\cite{ComputerVision}, biology~\cite{Biology}, and larger-scale parameter tuning.


High-dimensional BO comes with two key inter-related challenges \cite{Talk}: Identifying ``low-dimensional structure'' in the high-dimensional function, and choosing an acquisition function that can efficiently be optimized.  There is an inherent tension between these goals, with richer forms of structure often leading to acquisition functions that are harder to optimize. 


\vspace*{-1ex}
\subsection{Related Work}
\vspace*{-1ex}

A recent overview of BO can be found in \cite{Sha16}.  Most BO algorithms can be posed as choosing the next point to maximize an {\em acquisition function}, which in turn depends on the current posterior of the function.  Popular choices include upper confidence bound (GP-UCB) ~\cite{citeulike:2561921}, probability of improvement (PI)~\cite{Jones1998}, expected improvement (EI)~\cite{Mockus1994},~\cite{DBLP:journals/corr/abs-1012-2599}, and (predictive) entropy search (ES) \cite{Hen12,Her14}.  In this paper, we are particularly interested in high-dimensional extensions of GP-UCB.

The earliest works on high-dimensional BO considered functions that only vary along a low-dimensional subspace \cite{Chen2012BO},~\cite{wang2013bayesian},~\cite{NIPS2013_5152}.  While such approaches can be effective, this assumption on the function is rather strong, and achieving the full potential high-dimensional BO requires moving to richer classes.  A promising alternative approach was recently proposed by Kandasamy {\em et al.} ~\cite{conf/icml/KandasamySP15}, who modeled the function by a sum of independent low-dimensional functions, each defined on a fixed subset of the underlying variables.  

Crucially, the work of \cite{conf/icml/KandasamySP15} assumed that these subsets are {\em disjoint}.  This constraint considerably simplifies the optimization of the acquisition function, but restricts the space of functions that can be modeled.  A generalization of this approach was proposed in \cite{pmlr-v51-li16e} that allows for possible rotations within each term of the decomposition.

Within the framework of \cite{conf/icml/KandasamySP15}, a crucial challenge is learning the underlying additive decomposition from samples (typically while simultaneously performing optimization).  In \cite{conf/icml/KandasamySP15}, this was done based on randomly sampling the decompositions and choosing one to maximize the likelihood.  More efficient approaches were proposed by Wang {\em et al.}~\cite{wang2017batched} based on Gibbs sampling, and by Gardner {\em et al.}~\cite{pmlr-v54-gardner17a} based on a simple Markov Chain Monte Carlo (MCMC) method.

Additive models, not necessarily making use of Gaussian processes, have also found extensive use in other contexts such as function learning, e.g., see \cite{ravikumar2007spam,tyagi17algorithms} and the references therein.

\vspace*{-1ex}
\subsection{Contributions}
\vspace*{-1ex}

The main contributions of this paper are as follows:
\vspace*{-1.5ex}
\begin{itemize}
\setlength\itemsep{0ex}
\item We generalize the additive model of \cite{conf/icml/KandasamySP15} by allowing the additive model to consist of functions defined on {\em general} subsets of the underlying variables that need not be disjoint.  
\item By representing the interactions between the groups (i.e., the subsets on which the low-dimensional functions are defined) using a graph, we deduce an efficient high-dimensional variant of GP-UCB based on message passing.
\item We present a Gibbs sampling algorithm for learning the structure of the graph from data, having a similar flavor to that of \cite{wang2017batched} while addressing new challenges for the case of overlapping groups.
\item We demonstrate the effectiveness of our approach on both synthetic and real data sets, including improved versatility compared to the work of \cite{conf/icml/KandasamySP15}.
\item In the supplementary material, we generalize certain aspects of the mathematical analysis from  \cite{conf/icml/KandasamySP15}, and discuss their possible implications towards providing regret bounds.
\end{itemize}

We very recently learned of a closely related independent parallel work \cite{hoang2018decentralized} adopting a similar model, but using an {\em approximate} decentralized approach to optimize the acquisition function, and learning the graph via an alternative approach building on \cite{pmlr-v54-gardner17a}.

\section{Generalized Additive GP Model}

We consider the optimization of a $D$-dimensional function $f(x)$, where $x = (x_1,\dotsc,x_D)$ is the input vector.  We focus on discrete domains, where each variable $x_i$ takes values in some finite set $\Xc_i$ (though this set could represent the quantization of a real interval such as $[0,1]$).  Hence, the high-dimensional domain is $\Xc = \Xc_1 \times \dotsc \times \Xc_D$.  Each time we query the function $f$ at some point $x \in \chi$, we obtain a noisy observation $y = f(x) + \epsilon$ where $\epsilon \sim \mathcal{N}(0,\,\eta^{2})$. We aim at maximizing this function over the domain $\Xc$, i.e. finding $x_{\mathrm{opt}} = \argmax_{x\in \Xc} f(x)$.

\paragraph{Additive structure:} Following the work of Kandasamy {\em et al.}~\cite{conf/icml/KandasamySP15}, we assume that the target function can be decomposed into a sum of low-dimensional components as follows:
\begin{equation}
f(x) = f^{(1)}(x^{(1)}) + f^{(2)}(x^{(2)}) + ... + f^{(M)}(x^{(M)}),
\label{fdecomp}
\end{equation}
where each $x^{(i)} \in \Xc^{(i)} \subseteq \Xc$ is a low-dimensional component, with $\Xc^{(i)}$ being the product of a small number of $\Xc_k$.  The variables involved in this product are referred to as the {\em $i$-th group}, and this set of variables is denoted by $\Gc^{(i)} \subseteq \{1,\dotsc,D\}$.

In~\cite{conf/icml/KandasamySP15}, it was assumed that the different variable sets $\Gc^{(i)}$ do not overlap, i.e. $\Gc^{(i)} \cap \Gc^{(j)} = \emptyset$ for all $(i,j)$. In our setting, we allow for arbitrary overlaps between these variable sets, thereby permitting a significantly richer model class that is suited to \emph{interacting} groups.

{\bf Prior and posterior:} We assume that each term $f^{(i)}$ is an independent sample from a Gaussian process $\GP(\mu^{(i)}, \kappa^{(i)})$. As a result, the overall target function is also a sample from a $GP$: $f\sim \GP(\mu, \kappa)$ where
\begin{align}
\mu(x) &= \sum_{i=1}^M \mu^{(i)}(x^{(i)}) \\
\kappa(x, x') &= \sum_{i=1}^M \kappa^{(i)}(x^{(i)}, x'^{(i)})
\end{align}

Let $\mathcal{D}_t = \{(x_i, y_i)\}_{i=1}^t$ be the data observed from the target function $f$ where $\yv=(y_1,...,y_n)$ are the noisy observations corresponding to $\xv=(x_1,...,x_n)$, i.e. $y_i \sim \mathcal{N}(f(x_i), \eta^2), i=1,...,n$. Conditioned on these observations $\mathcal{D}_t$, we can infer the posterior mean and variance for each term $f^{(i)}$ at an arbitrary point $x_*$.  We show in the supplementary material that $(f_*^{(j)} | \yv) \sim \mathcal{N}(\mu_{t-1}^{(i)}, (\sigma_{t-1}^{(i)})^2)$, where
\begin{equation}
\begin{split}
\mu_{t-1}^{(j)}
&= \kappa^{(j)}(x_*^{(j)}, \xv^{(j)})\Delta^{-1}\yv \\
(\sigma_{t-1}^{(j)})^2
&= \kappa^{(j)}(x_*^{(j)}, x_*^{(j)}) \\
&\hspace*{0.2cm}- \kappa^{(j)}(x_*^{(j)},\xv^{(j)})\Delta^{-1}\kappa^{(j)}(\xv^{(j)}, x_*^{(j)})
\end{split}
\label{AddPosterior}
\end{equation}
with $\Delta = \kappa(\xv,\xv) + \eta^2I_n \in \mathbb{R}^{n\times n}$.
Here $\kappa ^{(j)}(\xv^{(j)}, x_*^{(j)})$ is a column vector of size $n$ whose $i$-{th} entry is $\kappa ^{(j)}(x_i^{(j)}, x_*^{(j)})$, and $\kappa (\xv,\xv)$ is a matrix of size $n\times n$ whose $(i,i')$-{th} entry is $\kappa(x_i, x_{i'})$.

\begin{figure}
\centering
\begin{tikzpicture}[scale=0.6, shape=circle, minimum size = .75cm, thick]
\node[draw, minimum size=0.7cm] (1) at (0,0) {1};
\foreach \phi in {2,...,6}{
\node[draw, minimum size=0.7cm] (\phi) at (360/5 * \phi:2.2cm) {$\phi$};
}
\path[draw,thick,line width=1.5pt] (1) edge node {} (2);
\path[draw,thick,line width=1.5pt] (1) edge node {} (3);
\path[draw,thick,line width=1.5pt] (2) edge node {} (3);
\path[draw,thick,line width=1.5pt] (1) edge node {} (4);
\path[draw,thick,line width=1.5pt] (3) edge node {} (4);
\path[draw,thick,line width=1.5pt] (4) edge node {} (5);
\end{tikzpicture}
\par
\caption{Example dependency graph. There are $4$ maximal cliques: (1,2,3), (1,3,4), (4,5), and (6), each of which are associated to a term of the decomposition.} \vspace*{-2ex}
\label{depGraph}
\end{figure}
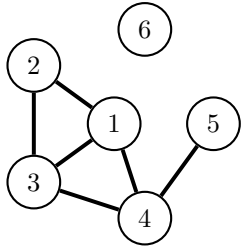

{\bf Dependency graph:} An additive decomposition can be represented by a {\em dependency graph}. The dependency graph is built by joining variables $i$ and $j$ with an edge whenever they appear together within some set $x^{(k)}$. For example, the graph associated with the decomposition of $f(x)=f^{(1)}(x_1,x_2,x_3) + f^{(2)}(x_1,x_3,x_4) + f^{(3)}(x_4,x_5) + f^{(4)}(x_6)$ is shown in Figure~\ref{depGraph}.  In the special case of \cite{conf/icml/KandasamySP15}, the dependency graph consists of disjoint fully connected components.


While different decompositions may lead to the same graph (e.g., consider the case of three functions on $(x_1,x_2)$, $(x_2,x_3)$, and $(x_1,x_3)$ vs.~a single function on $(x_1,x_2,x_3)$), one can always adopt the more general decomposition for a given graph to avoid any loss of generality (e.g., take the single function on $(x_1,x_2,x_3)$).  To this end, one can let each low-dimensional function correspond to a single {\em maximal clique} (i.e., fully-connected component) of the graph.

\vspace*{-1ex}
\section{Generalized Additive GP-UCB} \label{sec:alg}
\vspace*{-1ex}

\subsection{Description of algorithm}

{\bf Acquisition function:} We assign to each term $f^{(i)}$ a low-dimensional UCB acquisition function~\cite{icml2010_SrinivasKKS10}:
\begin{equation}
\phi_t^{(i)}(x^{(i)}) = \mu_{t-1}(x^{(i)}) + \beta_t^{\frac{1}{2}}\sigma_{t-1}(x^{(i)}), \label{eq:phi_i}
\end{equation}
where $\beta_t$ is an exploration parameter. The global acquisition function is the sum of low-dimensional ones:
\begin{equation}
\phi_t(x) = \sum_{i=1}^M \phi_t^{(i)}(x^{(i)}). \label{eq:acq_func}
\end{equation}
The bulk of this section is devoted to methods for efficiently maximizing this function.  While this is straightforward for disjoint groups \cite{conf/icml/KandasamySP15}, it is non-trivial in our generalized model.

{\bf Full algorithm:}  Our algorithm, which we call {\em generalized additive GP-UCB} (G-Add-GP-UCB), is given in Algorithm \ref{FullAlgo}.  As stated, the algorithm is suited to the case that the kernel (and hence the dependency graph) is known; this assumption is dropped in Section \ref{sec:learn_graph}.

\begin{algorithm}
\begin{algorithmic}[1]
\STATE Pick $(x_t)_{i=1}^{N_{\mathrm{init}}}$ at random; evaluate $f$ at these points to get $(y_t)_{i=1}^{N_{\mathrm{init}}}$ and add these pairs to $D_0$.
\FOR{ $t=1,...,N_{\mathrm{iter}}$}{
\STATE Perform Bayesian posterior updates conditioned on $D_{t-1}$ to obtain $\mu_{t-1}^{(j)}$, $\sigma_{t-1}^{(j)}$ for $j = 1,...,M$
\STATE Optimize the acquisition function ({\em cf.}, Section \ref{sec:opt_acq}) to obtain $x_t \leftarrow \argmax_{x\in \chi} \tilde{\phi}_t(x)$
\STATE Evaluate $x_t$ and observe $y_t \leftarrow f(x_t) + \epsilon$\;
\STATE Augment the data set : $D_t = D_{t-1}\cup{(x_t,y_t)}$
}
\ENDFOR
\end{algorithmic}
\caption{G-Add-GP-UCB \label{FullAlgo}}
\end{algorithm}

\subsection{Maximizing the acquisition function} \label{sec:opt_acq}

The key to maximizing the acquisition function (when the dependency graph is known) is to connect the optimization with the problem of maximizing probability in Markov random fields \cite{Wainwright2015}.  In particular, in the same way as the latter setting, we can make use of message passing for efficient maximization.  We first explain how this is done when the dependency graph is triangulated (i.e., when the graph has no chordless cycles of length greater than 3), and then extend the algorithm to general graphs.

{\bf Triangulated dependency graphs:} We start by constructing a junction tree \cite{Wainwright2015} of the dependency graph. The nodes of the junction tree are then the different maximal cliques of the dependency graph. By the construction of the dependency graph in the previous section, there is a low-dimensional acquisition function $\phi^{(C)}$ of the form \eqref{eq:phi_i} associated to each of these maximal cliques $C$.

Starting from the leaves of the junction tree, we sequentially maximize, for each clique $C$, the associated low-dimensional acquisition functions $\phi^{(C)}$, and pass ``messages'' $m_{C_p\leftarrow C}$ to the parent node $C_p$.  Each message is a function of the variables of the parent node, with the variables of the previous nodes already optimized. This process propagates up to the root, with a given node adding the messages from its children ({\em cf.}, Algorithm~\ref{Message Passing algorithm}). The running intersection property of the junction tree ensures that there is no conflict between the maximizations performed, and that this algorithm returns the maximum of the global acquisition function \cite{Wainwright2015}.

\begin{algorithm}
\begin{algorithmic}[1]
\STATE Root the junction tree at a random node $C_R$.
\STATE $d \leftarrow $ depth of rooted tree G.
\WHILE{$d \geq 1$}{
\FOR{each node $C$ of the junction tree at distance $d$ from the root}{
\STATE $C_p \leftarrow \text{parent(C)}$, $I \leftarrow C\cap C_p$, $J \leftarrow C\setminus C_p$
\STATE Compute messages to be passed to the parent node: \\
$m_{C_p\leftarrow C}(x^{(I)}) = \max_{x^{(J)}} \phi^{C}(x^{(C)}) + \sum_{C_c\in \text{Children}(C)}m_{C \leftarrow C_c}(x^{(C\cap C_c)})$ \\
for each $x^{(I)}\in \chi^{(C)}$
}
\ENDFOR
\STATE $d \leftarrow d-1$.
}
\ENDWHILE
\STATE Return $\max_{x^{(C_R)}} \phi^{(C_R)}(x^{(C_R)}) + \sum_{C_c\in \text{Children}(C_R)} m_{C_R\leftarrow C_c}(x^{(C_R\cap C_c)})$.
\end{algorithmic}
\caption{Message Passing algorithm on a triangulated dependency graph}
\label{Message Passing algorithm}
\end{algorithm}

{\bf Arbitrary dependency graphs:} When the dependency graph is not triangulated, there is no junction tree that can be constructed from the original dependency graph. We thus first need to triangulate it, and construct a junction tree for the processed dependency graph. However, since the dependency graph has been modified, we no longer directly have a unique low-dimensional acquisition function for each clique of the junction tree.  In order to ensure that each function $\phi^{(i)}$ is maximized once and only once during the optimization process, we assign to each clique $C$ the following ``acquisition function'':
\begin{equation}
\phi^{(C)}(x^{(C)}) = \sum\limits_{\substack{\textrm{clique}\ c\ of\ G,\ c\subset C, \\ c \notin C_c\ \forall C_c\in \textrm{Children}(C)}} \phi^{(c)}(x^{(c)}),
\label{JTfunctions}
\end{equation}
where $G$ corresponds to the original dependency graph. By doing so, we can then apply Algorithm~\ref{Message Passing algorithm} on the triangulated graph.

{\bf Complexity:} The complexity of running the message-passing algorithm on a junction tree $J$ is exponential in the size of the maximum clique of the triangulated graph associated with junction tree $J$. This quantity depends on the chosen triangulation, so it would be desirable to compute a triangulation that yields a small maximal clique \cite{arnborg1987complexity,cano1995heuristic}.


\vspace*{-1ex}
\section{Learning the Dependency Graph} \label{sec:learn_graph}
\vspace*{-1ex}

One of the main important practical challenges of BO is choosing a suitable kernel.  In the high-dimensional setting, this challenge is even more difficult, as we need to learn not only the kernel parameters (e.g., length scales), but also the structure associated with the high-dimensional function (i.e., the dependency graph).  In this section, we present a Gibbs sampling procedure for this purpose, building on the approach of \cite{wang2017batched} for the case of non-overlapping groups.


{\bf Preliminaries:} As discussed previously, any decomposition can be represented by an undirected graph, in which each low-dimensional kernel $\kappa^{(j)}$ is associated to a \textit{maximal clique}. For convenience, here we represent this decomposition by an adjacency matrix $Z \in \{0,1\}^{D \times D}$, where $Z_{ij} = 1$ if variable $i$ is connected to $j$ in the graph, and $0$ otherwise. This matrix is symmetric and has zeros on the diagonal, so the number of free parameters is $\frac{D(D-1)}{2}$.

The dependency graph defines the kernel decomposition, and thus influences the number of low-dimensional kernels and their dimensions. Therefore, in general, the kernel parameters can be defined only once the decomposition is known. In order to learn these parameters simultaneously with the additive structure, we assume that these kernels only depend on a constant number $n_{\mathrm{param}}$ of parameters {\em independent of the decomposition} (see Section \ref{sec:experiments} for a concrete example).  We then group them into the set $L = \{L_i\}_{i=1,...,n_{\mathrm{param}}}$. Overall, the parameters that must be optimized are $\theta = \{\{Z_{ij}\}_{1\leq i < j\leq D}, \{L_j\}_{1\leq j\leq n_{\mathrm{param}}}\}$, which results in $\frac{D(D-1)}{2} + n_{\mathrm{param}}$ parameters.

{\bf Maximum likelihood:} In order to infer the structure of the dependency graph and the kernels' parameters, we seek to maximize data likelihood:
\begin{align*}
\log{p(D_n | Z, L)} &= -\frac{1}{2}y^T\left(K_n^{G(Z), L} + \sigma^2I\right)^{-1}y \\
&-\frac{1}{2}\log{\left|K_n^{G(Z), L} +  \sigma^2I\right|} - \frac{n}{2}\log{2\pi},
\end{align*}
where $G(Z)$ is the graph corresponding to the adjacency matrix $Z$, $y$ is the vector containing the current observations, and $K_n^{G, L} \in \mathbb{R}^{n\times n}$ is the kernel matrix of the observed data points, supposing a dependency graph $G$ and kernel parameters $\{L_i\}_{1\leq i\leq n_{param}}$, i.e., $(K_n^{G, L})_{ij} = \kappa^{G,L}(x_i, x_j)$. Here $\kappa^{G,L}$ represents the high-dimensional kernel determined by $G$ and $L$.


{\bf Gibbs sampling:} We adopt a Bayesian approach, in which we place a prior distribution on the parameters $\{\theta_i\}_{1\leq i\leq N}$, and seek to sample from the posterior distribution $p(\theta_1, ..., \theta_N | D)$. Since we cannot directly sample from this high-dimensional probability distribution, we use the Gibbs sampling method ({\em cf.}, Algorithm \ref{Gibbs Sampling}).  This provides a means for approximately sampling from the joint distribution, as long as we can sample from the 1-dimensional conditional distributions $p(\theta_i | \theta_{-i}, D)$ with $\theta_{-i} = \{\theta_j | j\neq i\}$.

The algorithm starts from a set of parameters $\theta^{(0)}$, and iteratively samples new sets of parameters $\theta^{(j)}$ by modifying one coordinate at a time based on the posterior (Algorithm~\ref{Gibbs Sampling}). Once the sampling is performed, we choose the set of parameters that achieves the highest data likelihood.  It can be shown~\cite{ROBERTS1994207} that under soft assumptions on the probability distribution $p(\theta | D)$, the Gibbs sampler tends to sample the same way as if we were to sample directly from $p(\theta | D)$.

\begin{algorithm}
\begin{algorithmic}[1]
\STATE $\Theta \leftarrow \{\theta^{(0)}\}$, $\theta \leftarrow \theta^{(0)}$, $j \leftarrow 0$
\FOR{$j=1,2,\dotsc$}{
\FOR{$i=1,...,N$}{
\STATE $\theta^{(j+1)} \leftarrow \theta^{(j)}$.
\STATE Sample $\theta_i^{\mathrm{new}}$ from $p(\theta_i | \theta^{(j)}_{-i}, D)$
\STATE $\theta^{(j+1)}_i \leftarrow \theta_i^{\mathrm{new}}$
\STATE Augment the data set : $\Theta \leftarrow \Theta \cup \{\theta^{(j+1)}\}$
}
\ENDFOR
}
\ENDFOR
\end{algorithmic}
\caption{Structure learning via Gibbs Sampling}
\label{Gibbs Sampling}
\end{algorithm}

{\bf Prior distributions:}  We model the variables $\{Z_{ij}\}_{1\leq i < j\leq D}$ as Bernoulli random variables with parameter $p$: $Z_{ij} \sim \Bernoulli(p)$ where $p$ gives the probability of an edge joining variables $i$ and $j$. This parameter $p$ can be used to control the sparsity of the graph, or set to $\frac{1}{2}$ if no prior information is available.


Using this model, the posterior distribution for $Z_{ij}$ is
\begin{align*}
& p(Z_{ij} = 1 | Z_{-(ij)}, L, D_n; p) \\
&\propto p(D_n | Z_{-(ij)}, Z_{ij} = 1, L)\cdot p(Z_{ij}=1 | Z_{-(ij)}, L; p) \\
&= p(D_n | Z_{-(ij)}, Z_{ij} = 1, L) \cdot p \\
&\propto p\cdot e^{\phi(Z_{-(ij)}\cup Z_{ij} = 1, L)}
\end{align*}
with $\phi(Z,L) = -\frac{1}{2}y^T(K_n^{G(Z), L} + \sigma^2I)^{-1}y - \frac{1}{2}\log{\big|K_n^{G(Z), L} +  \sigma^2I\big|}$.  Concerning the kernels' parameters, we simply model them as uniform variables over pre-defined sets of possible values: $L_{i} \sim \textrm{Uniform}(\mathcal{L}_{i})$ so that for each $l \in \mathcal{L}_{i}$, we have
\begin{equation}
p(L_{i} = l | Z, L_{-i}, D_n; p) \propto e^{\phi(Z, L_{-i}\cup \{L_{i}=l\})}
\end{equation}
Using these posterior distributions, we can apply Gibbs sampling to the parameter set $\theta = \{\{Z_{ij}\}_{1\leq i < j\leq D}, \{L_{i}\}_{1\leq i \leq n_{param}}\}$, and select the set which produces the highest data likelihood.  For the binary variables $Z_{ij}$, we simply compute $p_0 = (1-p)\cdot e^{\phi(Z_{-(ij)}\cup \{Z_{ij} = 0\}, L)}$ and $p_1 = p\cdot e^{\phi(Z_{-(ij)}\cup \{Z_{ij} = 0\}, L)}$ and then sample from the binary distribution $\Bernoulli(\frac{p_1}{p_0+p_1})$. For the kernels' parameters, we proceed the same way by computing $e^{\phi(Z, L_{-i}\cup \{L_{i}=l\})}$ for values $l$ in the set $\mathcal{L}_{i}$ and then sample from the normalized probability distribution.

{\bf Stopping criterion:} We stop the process after some number $N_{\mathrm{Gibbs}}$ of data likelihoods have been computed. The more data we have, the better the algorithm will perform to find the true dependency graph, or a closely related one. Therefore, this learning process is repeated throughout the Bayesian optimization algorithm every $N_{\mathrm{cyc}}$ iterations, for some choice of $N_{\mathrm{cyc}}$. Each time we learn new parameters (i.e., graph structure and kernel parameters), we start the sampling from the previously learned parameters. 


\vspace*{-1ex}
\section{Experiments} \label{sec:experiments}
\vspace*{-1ex}

In this section, we experimentally compare our generalized algorithm against the one of \cite{conf/icml/KandasamySP15}. We focus on squared exponential kernels for modeling the low-dimensional components of the target function:
\begin{align*}
\kappa^{(i), L^{(i)}}&(x^{(i)},x'^{(i)}) \\
&= \sigma^{(i)} \exp \left(-\frac{1}{2}(x^{(i)} - x'^{(i)})^T L^{(i)}(x^{(i)} - x'^{(i)})\right)
\end{align*}
The scales $\sigma^{(i)}$ are set to $\sigma^{(i)} = \frac{d_i}{\sum_j d_j}$, so that $\kappa(x,x) = 1$. The matrices $L^{(i)}$ are all diagonal and generated from one $D$-dimensional lengthscale vector $l$: $(L^{(i)})^{-1} = \textrm{diag}(l_{\chi^{(i)}})^2$ where $l_{\chi^{(i)}}$ is the vector containing the lengthscales of variables within $\chi^{(i)}$. This vector is then learned from the data via Gibbs sampling together with the dependency graph.

\subsection{Experiments on synthetic data}
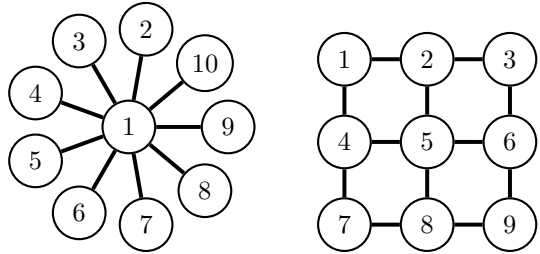
\begin{figure}
\begin{center}
\begin{tikzpicture}[scale=0.6, shape=circle, minimum size = .75cm, thick]
\node[draw, minimum size=0.7cm] (1) at (0,0) {1};
\foreach \phi in {2,...,9}{
\node[draw, minimum size=0.7cm] (\phi) at (360/9 * \phi:2.2cm) {$\phi$};
\path[draw,thick,line width=1.5pt] (1) edge node {} (\phi);
}
\node[draw, minimum size=0.7cm] (10) at (360/9 * 10:2.2cm) {$10$};
\path[draw,thick,line width=1.5pt] (1) edge node {} (10);
\end{tikzpicture}
\qquad 
\begin{tikzpicture}[scale=1.1, shape=circle, minimum size = .75cm, thick]
\node (1) at ( 0, 2) [draw, minimum size=0.7cm] {$1$};
\node (2) at ( 1, 2) [draw, minimum size=0.7cm] {$2$};
\node (3) at ( 2, 2) [draw, minimum size=0.7cm] {$3$};
\node (4) at ( 0, 1) [draw, minimum size=0.7cm] {$4$};
\node (5) at ( 1, 1) [draw, minimum size=0.7cm] {$5$};
\node (6) at ( 2, 1) [draw, minimum size=0.7cm] {$6$};
\node (7) at ( 0, 0) [draw, minimum size=0.7cm] {$7$};
\node (8) at ( 1, 0) [draw, minimum size=0.7cm] {$8$};
\node (9) at ( 2, 0) [draw, minimum size=0.7cm] {$9$};
\draw[line width=1.5pt] (1) -- (2);
\draw[line width=1.5pt] (2) -- (3);
\draw[line width=1.5pt] (1) -- (4);
\draw[line width=1.5pt] (2) -- (5);
\draw[line width=1.5pt] (3) -- (6);
\draw[line width=1.5pt] (4) -- (5);
\draw[line width=1.5pt] (5) -- (6);
\draw[line width=1.5pt] (4) -- (7);
\draw[line width=1.5pt] (5) -- (8);
\draw[line width=1.5pt] (6) -- (9);
\draw[line width=1.5pt] (7) -- (8);
\draw[line width=1.5pt] (8) -- (9);
\end{tikzpicture}
\end{center}
\caption{True dependency graphs for synthetic experiments. For both graphs, each edge forms a maximal clique, and thus corresponds to a term in the additive decomposition of the target function $f$. These graphs are not composed of disjoint cliques, and thus cannot be modeled by the constrained model without overlap.} \vspace*{-2ex}
\label{SyntheticDepGraphs}
\end{figure}

We first test our algorithm on synthetic data by sampling functions from Gaussian processes, according to the dependency graphs shown in Figure~\ref{SyntheticDepGraphs}. We use a squared exponential kernel with lengthscale matrix $L^{(i)} = 0.2 *I_2$. When applying Gibbs sampling use $p=\frac{1}{2}$ (i.e., no prior knowledge) for the sampling of the dependency graphs.

We apply our algorithm when considering 3 different cases, described as follows.

\paragraph{Overlap:}In this case, we learn both the lengthscales and the dependency graph, and allow for overlaps between the groups in the function decomposition. Thus, any dependency graph is allowed, and junction trees are used when optimizing the acquisition function.

\paragraph{No Overlap:} This case is similar to the previous one, except that we do not allow for overlaps between the groups in the function decomposition. This algorithm is thus the same as the one designed by Kandasamy {\em et al.} \cite{conf/icml/KandasamySP15}.  We place no ``hard constraints'' on the group sizes (i.e., restrictions on the maximum number of groups $M$ and maximum group size $\textrm{d}_{\textrm{max}}$ therein), but we apply the Gibbs sampling procedure of \cite{wang2017batched}, which only samples graphs that are sufficiently likely according to the samples.


\paragraph{Oracle:} in the last situation, we assume that we know the true dependency graph and lengthscales, and hence we do not need to learn them throughout the algorithm.  We expect this case to perform the best.

We also compare the results with a random algorithm that simply evaluates points at random. We perform 10 runs, starting from different initial situations where $10$ points are chosen at random (for each run, these randomly chosen points are the same for each model), and then run the simulation for a certain number of iterations. All the parameters are the same for each model, and are summarized in Table~\ref{tab:ParamSyntheticBO}.

\begin{table}[h!]
\centering
\begin{tabular}{c|c|c|c}
$\beta_t^{(i)}$ & $\textrm{N}_{\textrm{cyc}}$ & $\textrm{N}_{\textrm{Gibbs}}$ & $\textrm{max}_{\textrm{eval}}$ \\
\hline
$0.5\log{2t}$ & $30$ & $200$ & $1000$ \\
\end{tabular}
\caption{Parameters for the synthetic BO examples.} \vspace*{-2ex}
\label{tab:ParamSyntheticBO}
\end{table}

\begin{figure}[h!]
\centering
\begin{subfigure}{.4\textwidth}
\centering
\includegraphics[width=1.\linewidth]{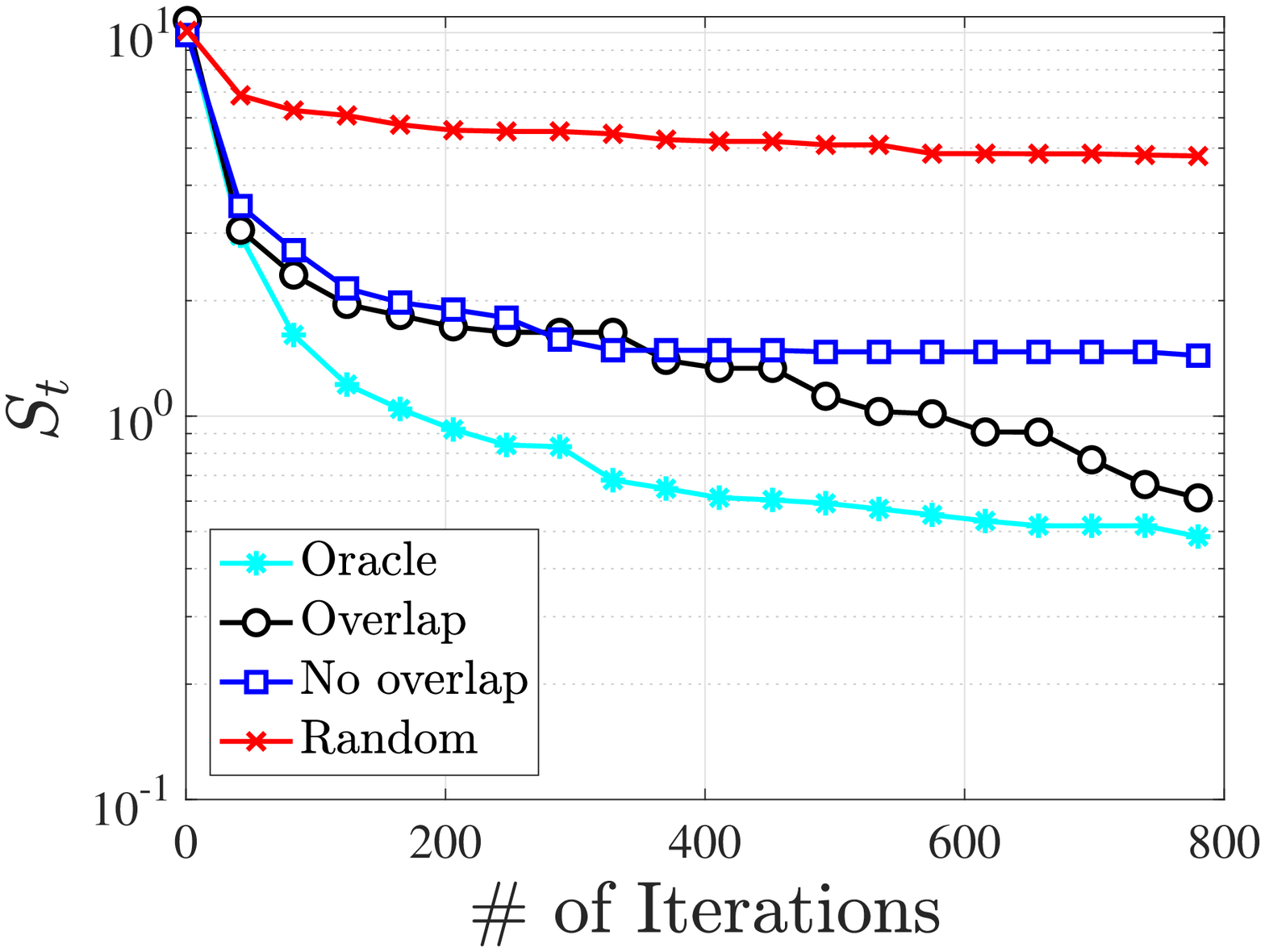}
\end{subfigure}
\begin{subfigure}{.4\textwidth}
\centering
\hspace*{2ex}\includegraphics[width=1.\linewidth]{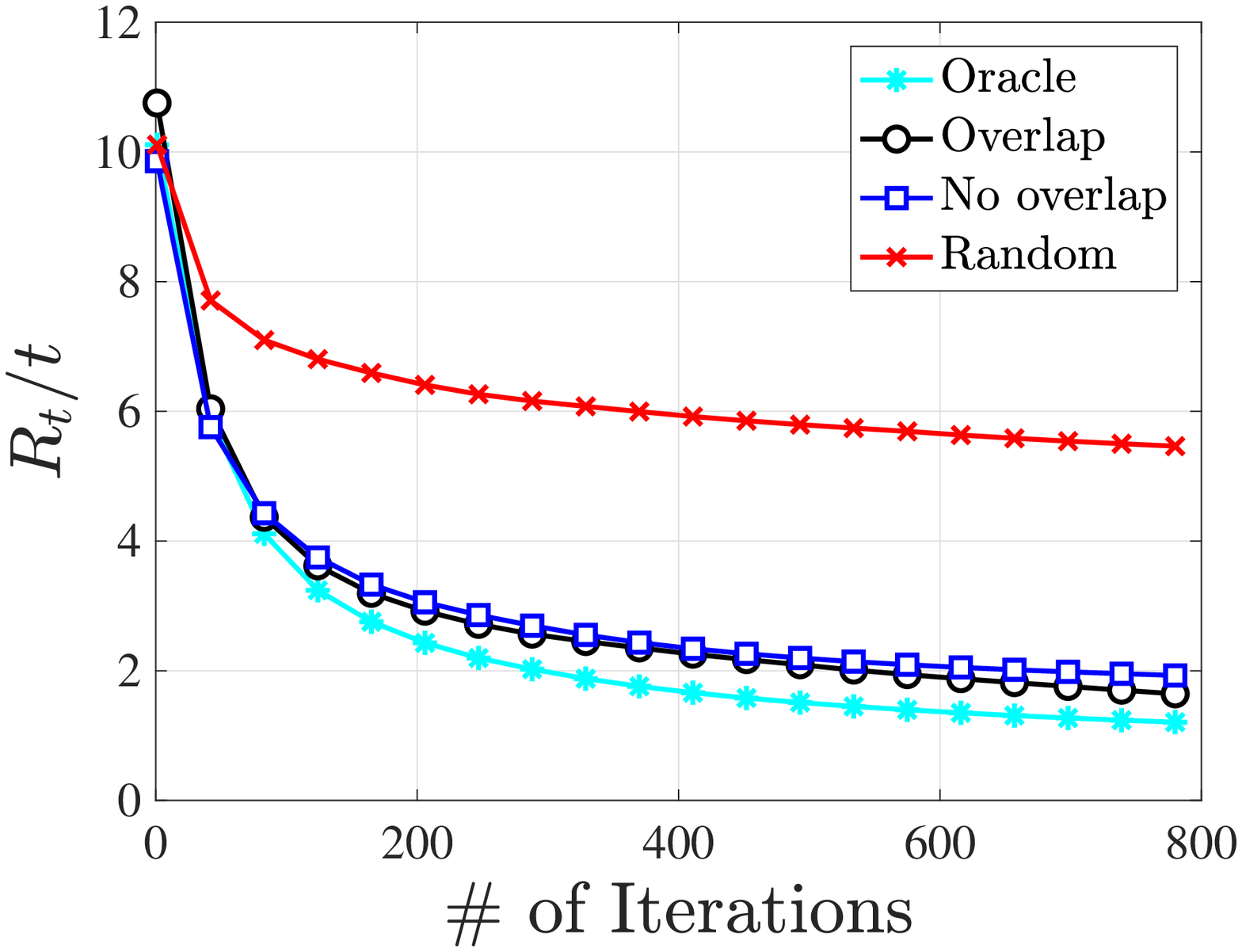}
\end{subfigure}
\caption{Star dependency graph: Simple and average cumulative regret averaged over $10$ runs.} \vspace*{-2ex}
\label{fig:RegretStar}
\end{figure}

\begin{figure}[h!]
\centering
\begin{subfigure}{.400\textwidth}
\centering
\includegraphics[width=1.\linewidth]{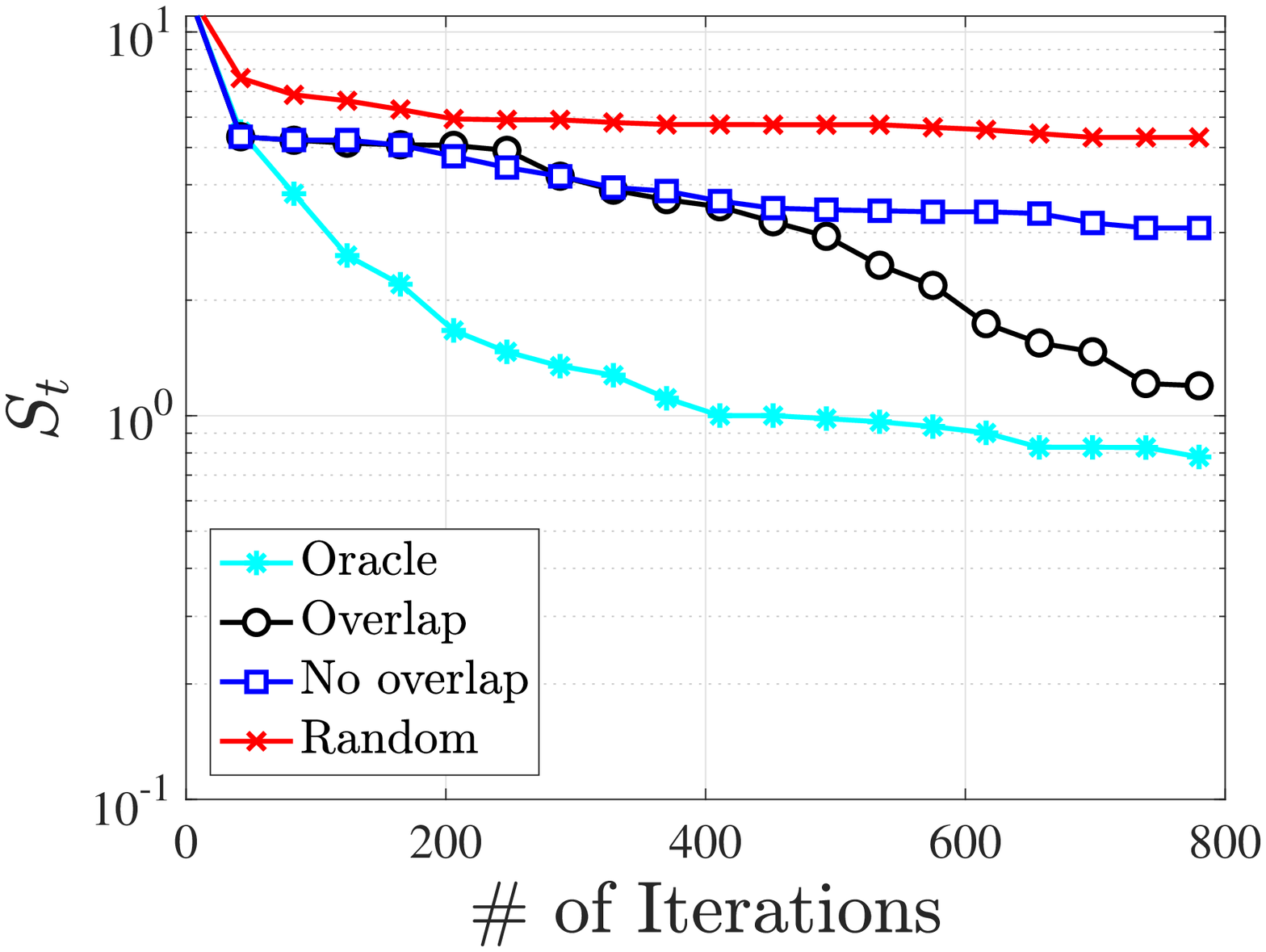}
\end{subfigure}
\begin{subfigure}{.4\textwidth}
\centering
\hspace*{2ex}\includegraphics[width=1.\linewidth]{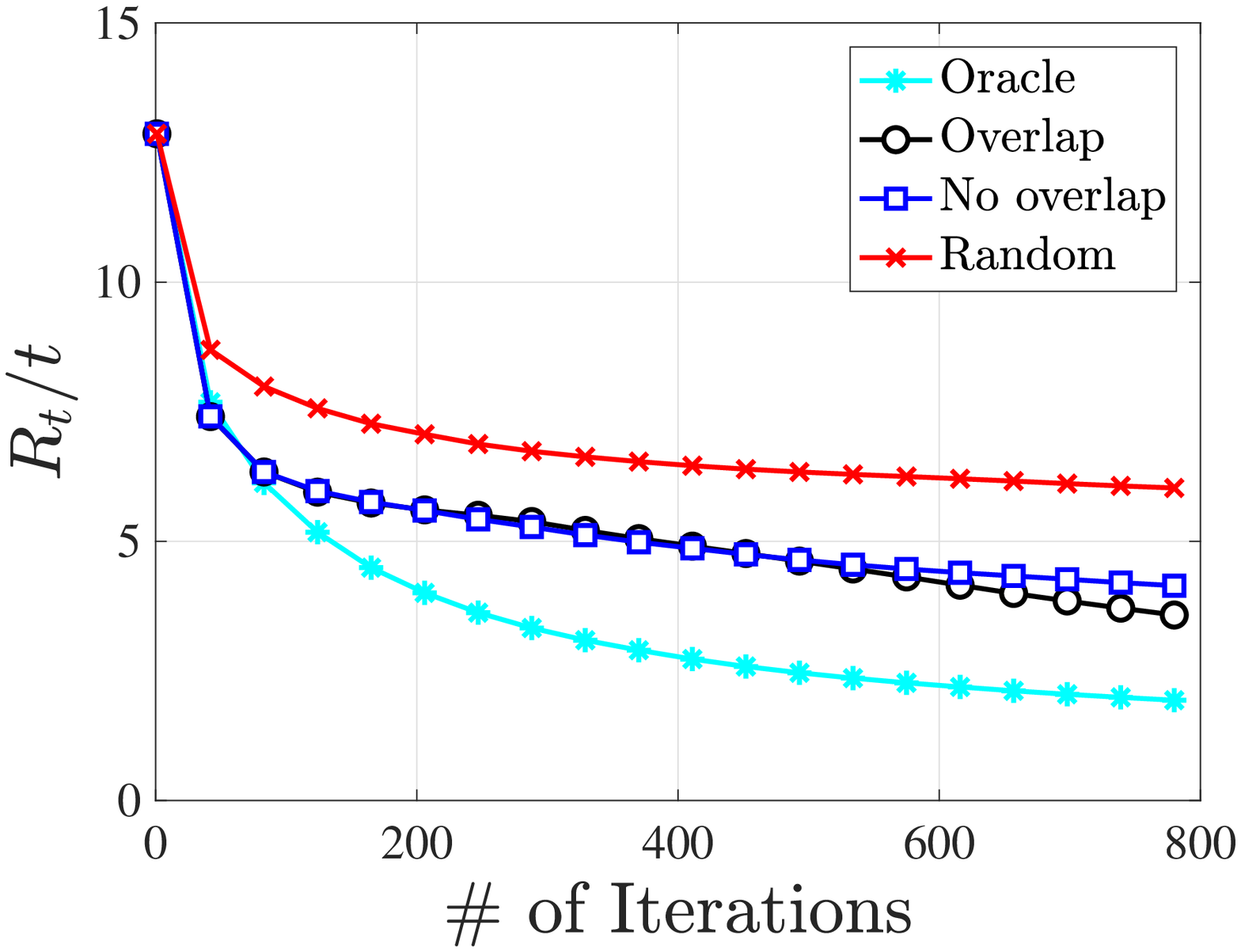}
\end{subfigure}
\caption{Grid dependency graph: Simple and average cumulative regret averaged over $10$ runs.} \vspace*{-2ex}
\label{fig:RegretGrid}
\end{figure}

\paragraph{Optimization performance:}
For each run and each iteration $t$, we compute the simple regret $S_t = \min_{i\leq t} r_i$ as well as the average cumulative regret $\frac{R_t}{t} = \frac{1}{t}\sum_{i=1}^t r_i$. We average these quantities over all trials (Figures~\ref{fig:RegretStar} and ~\ref{fig:RegretGrid}).

We observe that the knowledge of the variable dependencies indeed greatly improves the performance of the algorithm. Around the beginning of the algorithm, the learning process does not perform well as it uses too few data. Therefore, there is no significant difference between the models ``Overlap'' and ``No Overlap'', while the ``Oracle'' is much more efficient as it uses the true dependency graph.  As we get more data, the learning process improves and the ``Overlap'' model becomes more efficient compared to ``No Overlap''. 

\paragraph{Graph learning accuracy:}

It is also interesting to assess the learning of the dependency graph throughout the algorithm. To do so, we define two quantities in order to evaluate how close a graph $G$ is from a reference graph $G_{\mathrm{true}}$. We first define the Correct Connections quantity:
\begin{align*}
\textrm{CC}(G, &G_{\mathrm{true}}) = \frac{\text{\# edges in both G and $G_{\mathrm{true}}$}}{\text{\# edges in $G_{\mathrm{true}}$}},
\end{align*}
which describes how well the variables connections in the true graph are represented by the learned graph. This quantity is between $0$ and $1$, and is equal to $1$ if and only if all edges of the true graph are part of the edges of the learned graph. The second quantity that we define is the Correct Separation quantity:
\begin{align*}
&\textrm{CS}(G, G_{\mathrm{true}})
= \frac{\text{\# non-edges in both G and $G_{\mathrm{true}}$}}{\text{\# non-edges in $G_{true}$}}
\end{align*}
which describes how well the variables separations in the true graph are represented by the learned graph. This quantity is also between $0$ and $1$, and is equal to $1$ if and only if all edges of the learned graph are part of the edges of the true graph. It follows that $G_1 = G_2 \Leftrightarrow CC(G_1,G_2) = CS(G_1,G_2) = 1$.

\begin{figure}[h!]
\centering
\begin{subfigure}{.400\textwidth}
\centering
\includegraphics[width=1.\linewidth]{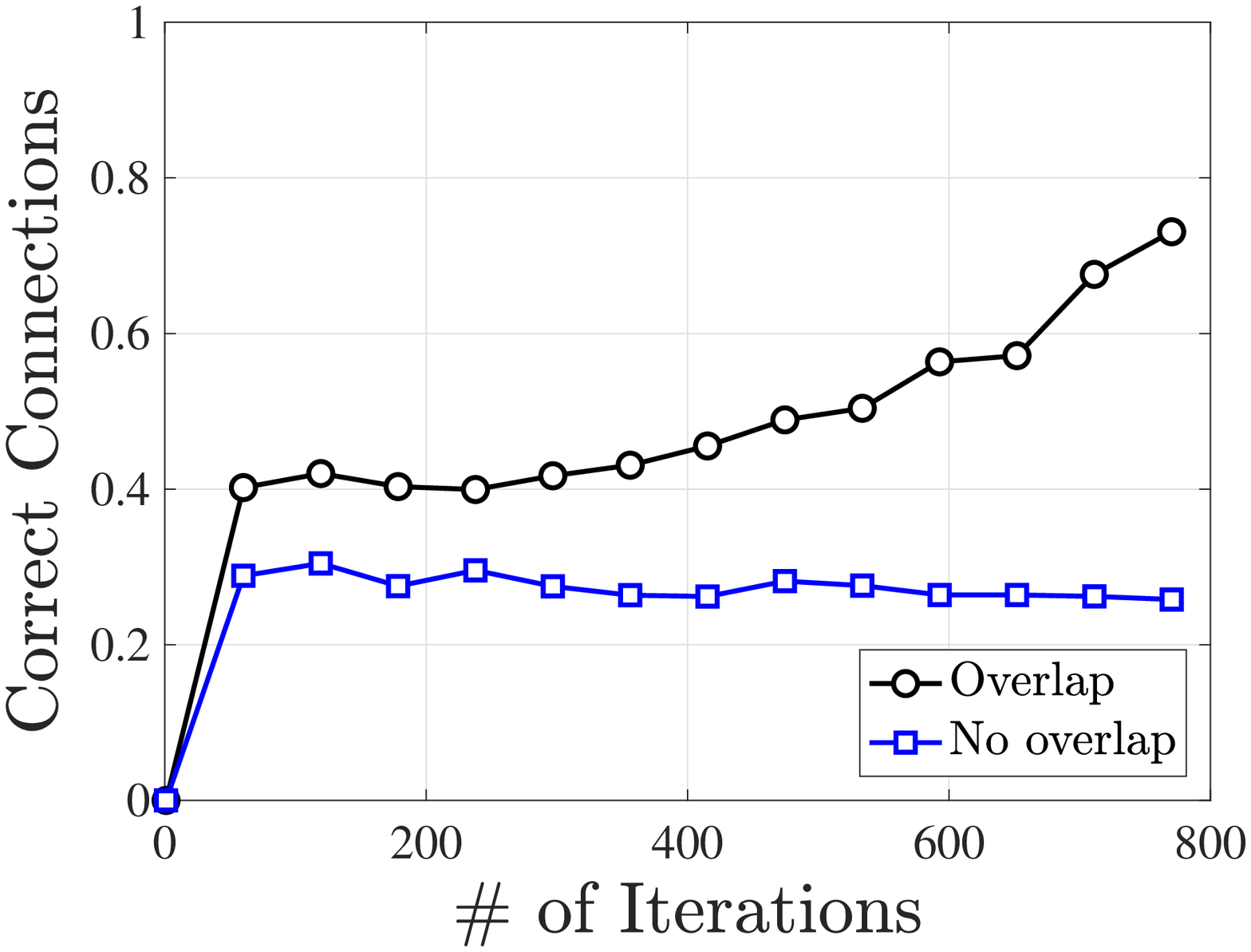}
\end{subfigure}
\begin{subfigure}{.400\textwidth}
\centering
\includegraphics[width=1.\linewidth]{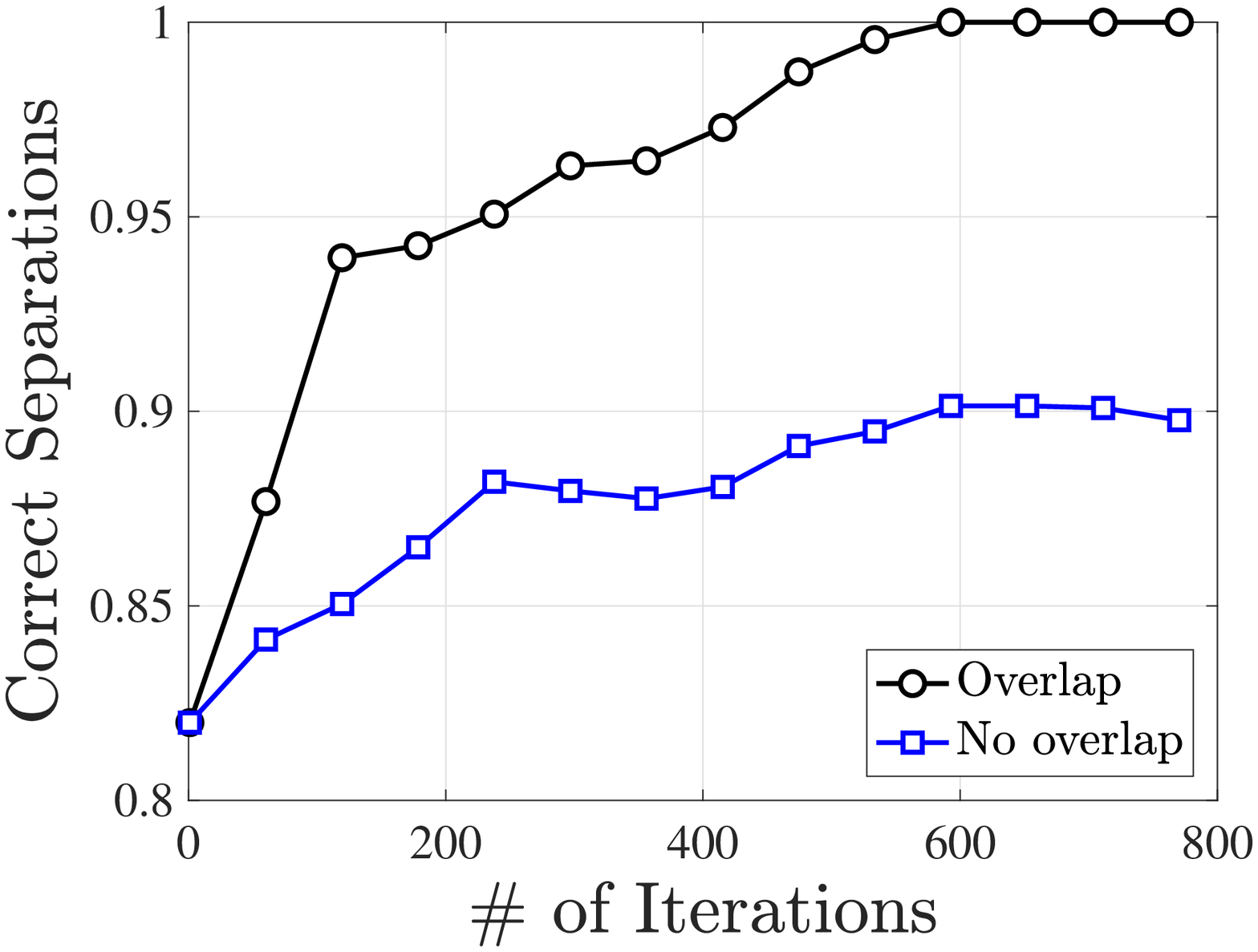}
\end{subfigure}
\caption{Star dependency graph: Correct Connections and Separations as a function of the number of data used for learning the graph.} \vspace*{-2ex}
\label{fig:GraphLearningStar}
\end{figure}

\begin{figure}[h!]
\centering
\begin{subfigure}{.400\textwidth}
\centering
\includegraphics[width=1.\linewidth]{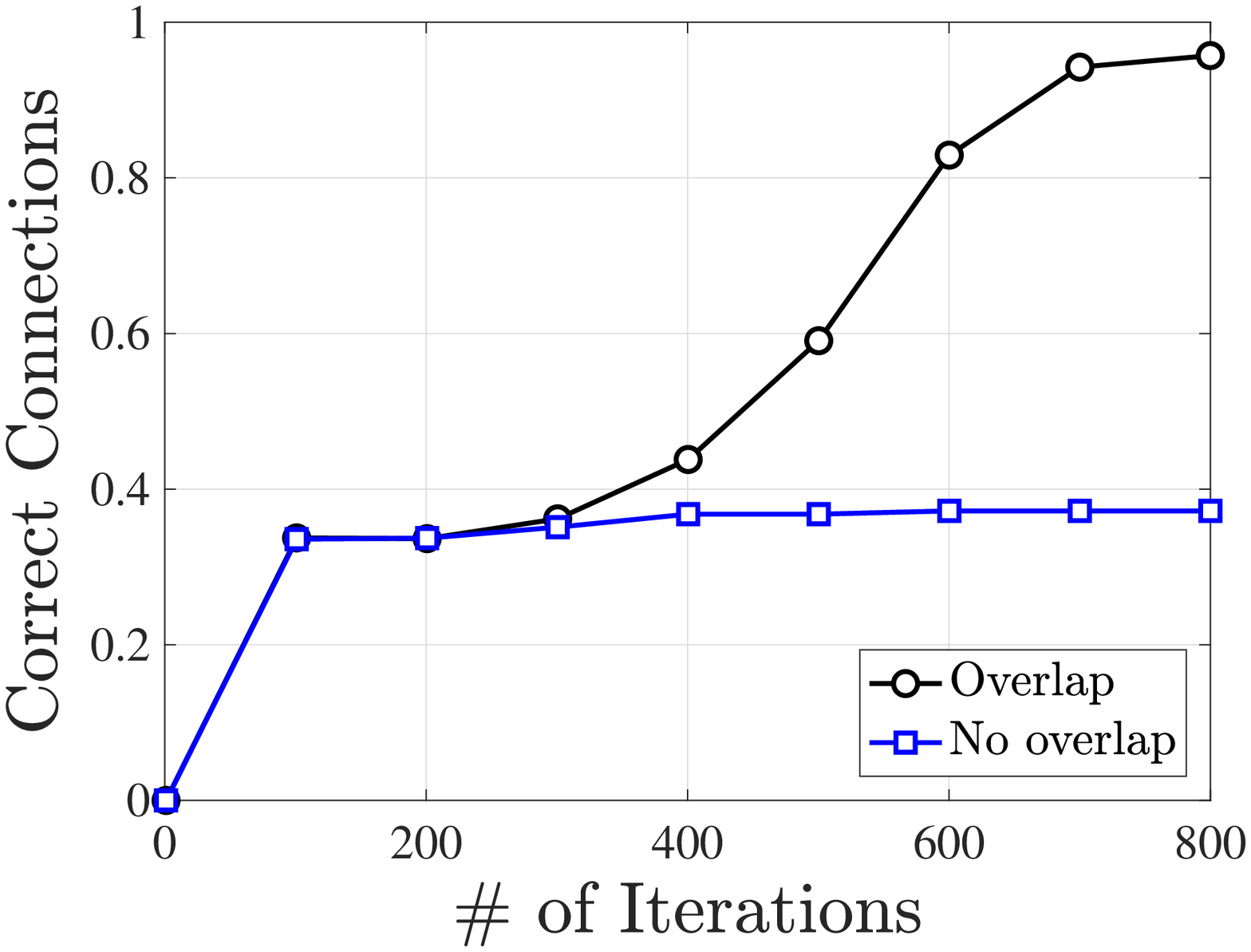}
\end{subfigure}
\begin{subfigure}{.400\textwidth}
\centering
\includegraphics[width=1.\linewidth]{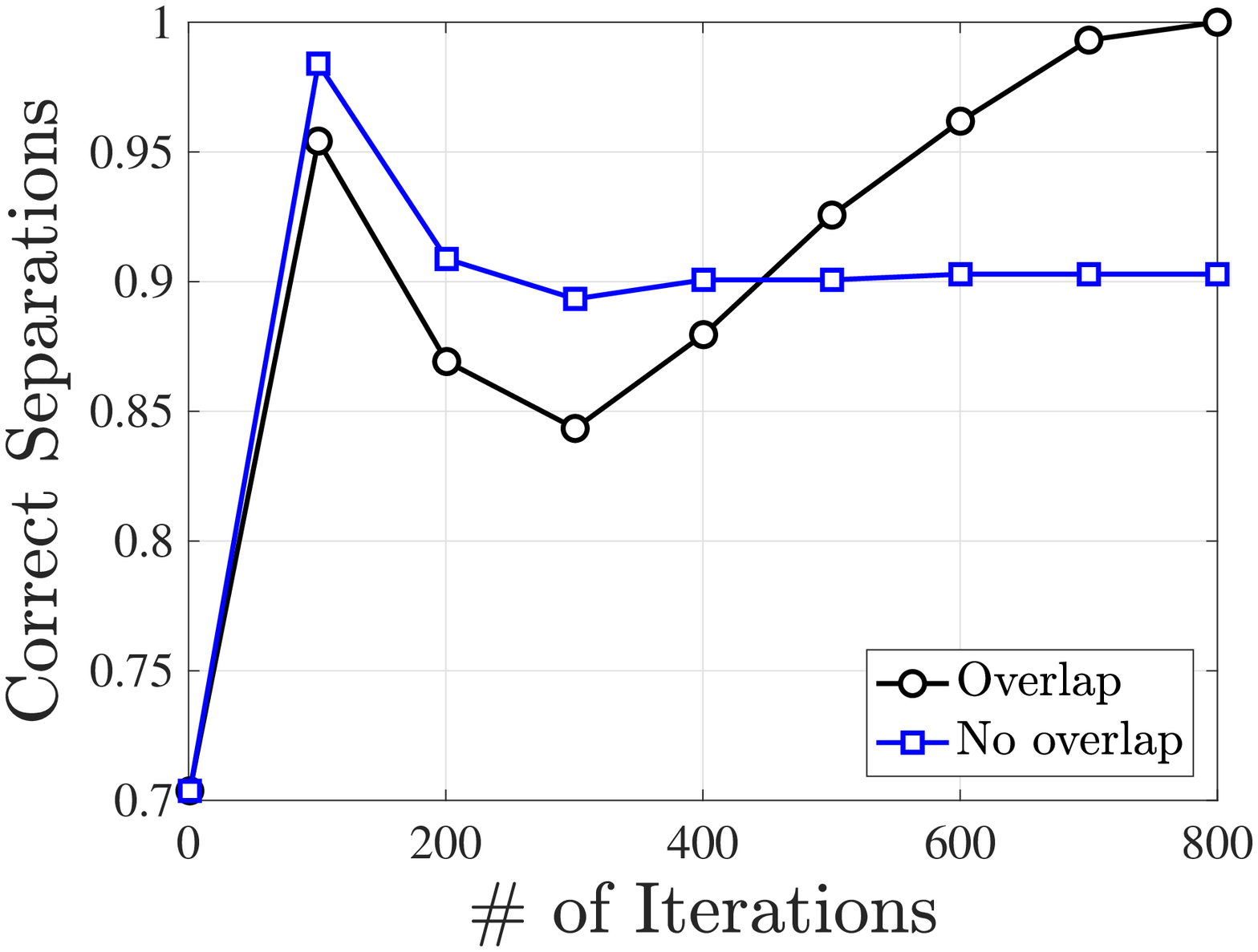}
\end{subfigure}
\caption{Grid dependency graph: Correct Connections and Separations as a function of the number of data used for learning the graph.} \vspace*{-2ex}
\label{fig:GraphLearningGrid}
\end{figure}

In Figures \ref{fig:GraphLearningStar} and \ref{fig:GraphLearningGrid}, we plot these quantities as we obtain more data, and observe that the learning process steadily improves over time. However, in the ``No Overlap'' case, the graph constraint induces a saturation in the learning, while in the ``Overlap'' case, we observe converges to the true graph.  We note that the high number of correct separations at early iterations is not an indicator of good performance, as the number of correct connections is very low.

We can observe that the time at which the ``Overlap'' model starts to be more efficient than the ``No Overlap'' one corresponds to the time where the dependency graph starts to be closer to the true graph.


\subsection{Experiments on real data}

\underline{Face recognition}: We consider the optimization of parameters for the Viola and Jones (VJ) Cascade Classifier~\cite{Viola01rapidobject}. This algorithm aims at recognizing faces in images. It is based on a cascade of increasingly complex classifiers, which all detect particular features on the image to classify. At each stage, the classifier searches on the image for some features. If it does not find it, the image is rejected, directly classified and no more processing is performed. Otherwise, it continues to the next stage, and so on. It is thus necessary that the first stages have a low rate of false negatives. 

Each classifier has a threshold parameter that controls this rejection. The problem is then to optimize these thresholds in order to obtain the highest possible classification accuracy. For each set of parameters, we evaluate the classification accuracy on $1500$ images. Among these images, $1000$ of them contain exactly one face, and the $500$ others do not contain any face. The algorithm correctly classifies an image containing a face if it detects exactly one face. The classification accuracy is defined as the proportion of correctly classified images. We use the OpenCV implementation of the Cascade VJ algorithm \cite{Bradski:2008:LOE:1461412}, which uses a $22$-stages cascade algorithm. OpenCV also provides an optimized set of thresholds, which gives a classification accuracy of $92.6\%$ on our data set. When applying our Bayesian optimization algorithm to this problem, we choose as the domain a neighborhood of the optimized parameter set given by OpenCV, and set the target function as the classification accuracy.

{\bf Methods:} As in the synthetic data example, we want to compare our ``Overlap'' model to the ``No Overlap'' one. In their paper \cite{conf/icml/KandasamySP15}, Kandasamy {\em et al.} applied their model to this same data set, and observed that the optimal extra parameters for this optimization problem were $M=4$ and $\textrm{d}_{\textrm{max}}=6$. We thus use these parameters for this model.

As discussed above, consecutive stages will have more similar rejection thresholds than stages that are far from each other. Therefore, we may suppose that two consecutive stages have more similar thresholds than stages that are far apart. This assumption is enforced by the fact that the set of parameters provided by OpenCV is increasing with the number of stages, justifying this possible correlation between close stages. Therefore, for our ``Overlap'' model, we set a dependency graph where each variable $i$ is connected to variables $i-1$ and $i+1$ $\forall i=2,...,21$, and do not apply the graph structure learning throughout the BO.

The Oracle method cannot be applied here since we do not know the true structure of the target function.

\begin{table}[h!]
\centering
\begin{tabular}{c|c|c|c|c}
$\textrm{N}_{\textrm{iter}}$ & $\beta_t^{(i)}$ & $\textrm{N}_{\textrm{cyc}}$ & $\textrm{N}_{\textrm{Gibbs}}$ & $\textrm{max}_{\textrm{eval}}$\\
\hline
$200$ & $0.5\log{2t}$ & $50$ & $300$ & $2000$\\
\end{tabular}
\caption{BO parameters for face recognition data.} \vspace*{-2ex}
\label{tab:ParamRealBO}
\end{table}

\begin{figure}[h!]
\centering
\begin{subfigure}{.400\textwidth}
\centering
\includegraphics[width=1.\linewidth]{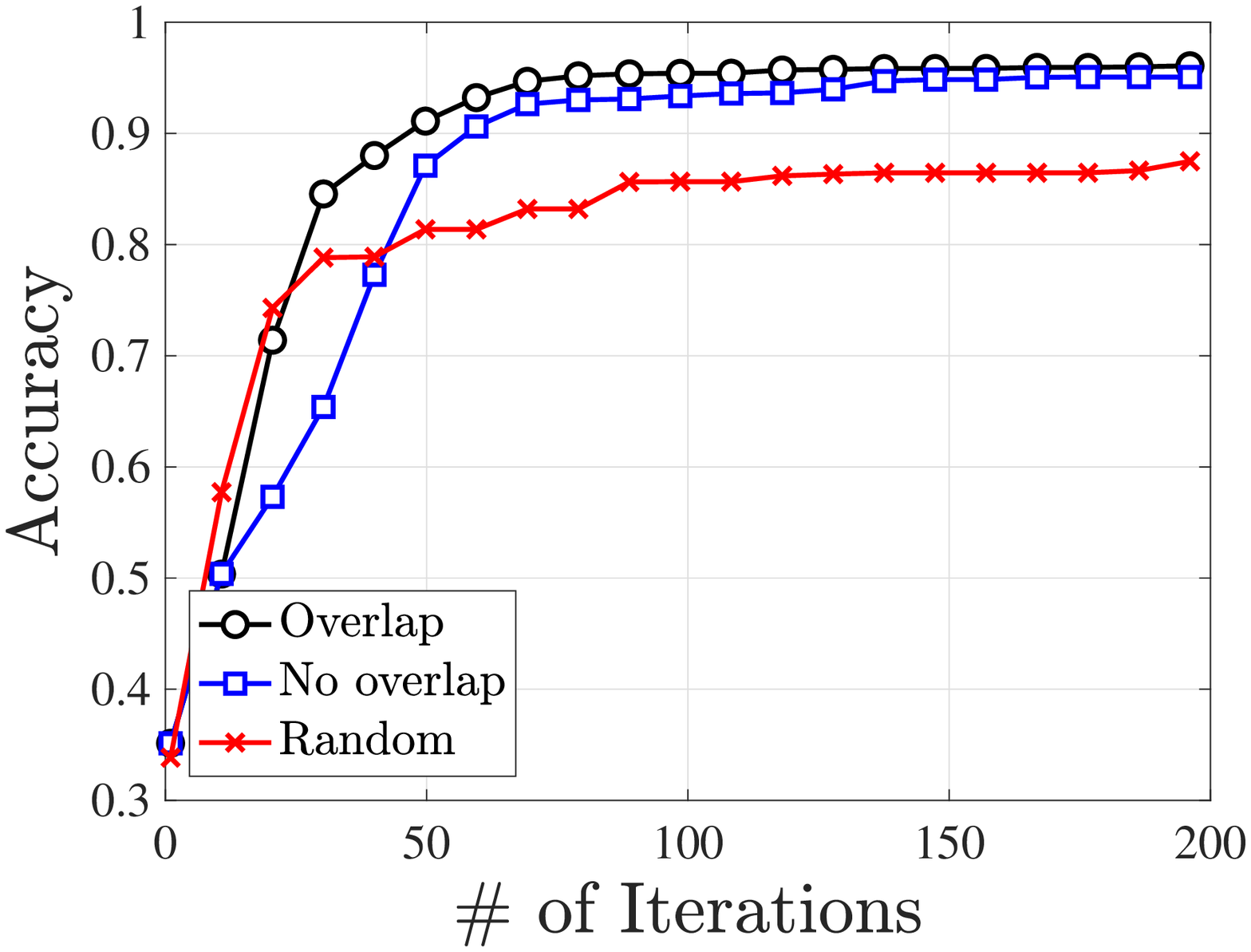}
\end{subfigure}
\begin{subfigure}{.400\textwidth}
\centering
\includegraphics[width=1.\linewidth]{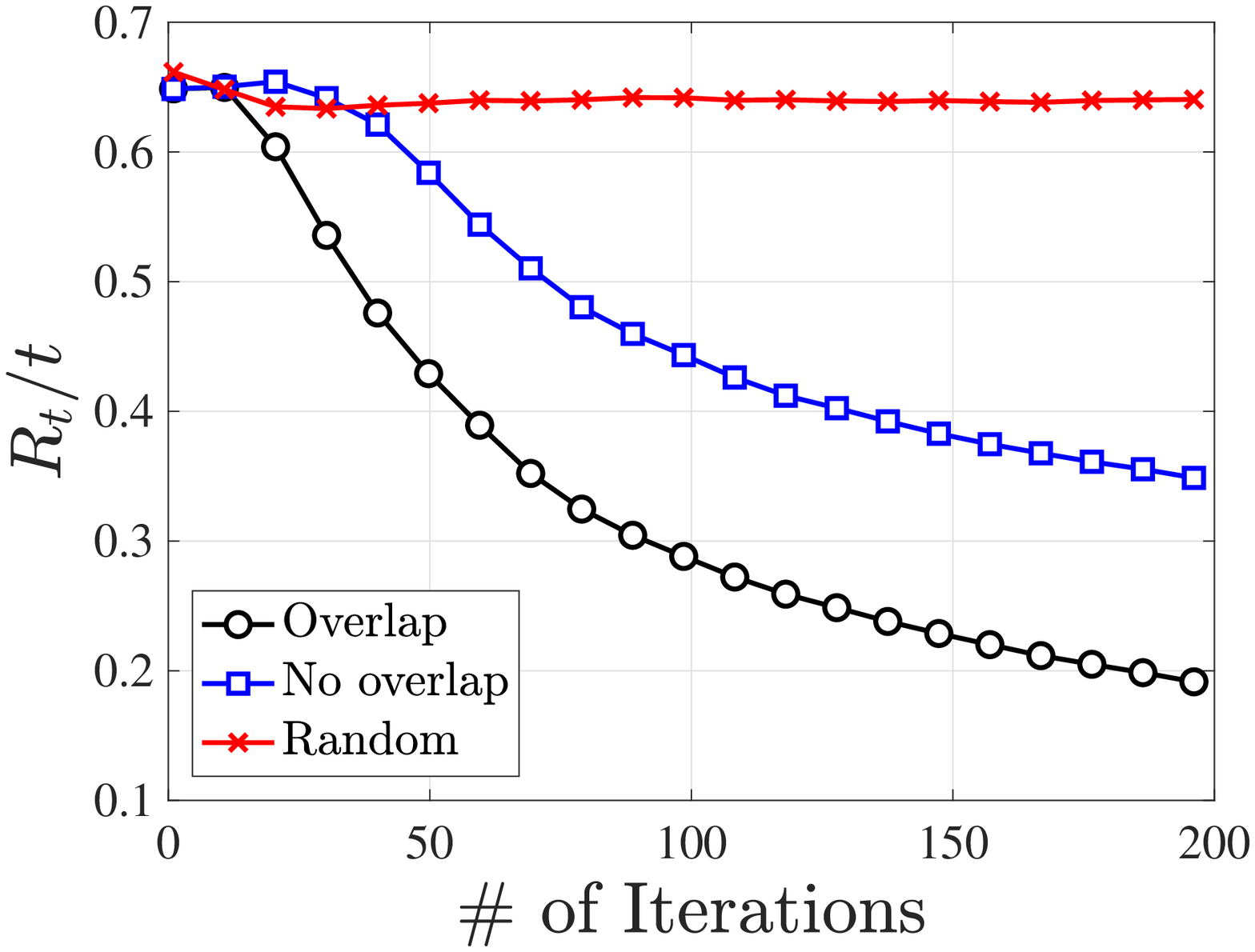}
\end{subfigure}
\caption{Face detection using VJ algorithm: classification accuracy and average cumulative regret.}
\label{fig:RealBOResults}
\end{figure}

{\bf Optimization performance:} We apply the two algorithms using the same set of parameters, described in Table~\ref{tab:ParamRealBO}. We also compare them to a random algorithm which queries random points on the same domain.  The results are given in Figure~\ref{fig:RealBOResults}, where we average over 15 runs.

We observe that the ``Overlap'' model performs slightly better than the ``No Overlap'' algorithm both in terms of convergence speed and optimal value. Both algorithms reach better classification accuracy than with the parameters provided by OpenCV. We can also observe that the ``Overlap'' model is much better in terms of the average cumulative regret.

\underline{Astrophysical data}: In the supplementary material, we provide a second example for real-world data based on maximizing likelihood in an astrophysical model.

\vspace*{-2ex}
\section{Conclusion}
\vspace*{-1ex}

We introduced a novel model and algorithm for high-dimensional Bayesian optimization, in which the target function to optimize can be represented by a sum of possibly overlapping low-dimensional components. By defining an acquisition function with the same additive structure, we can efficiently maximize it using message passing.  In addition, we proposed a Gibbs sampling method for learning the structure from samples. By expanding the space of functions that can be modeled, we observed through experiments that the efficiency of Bayesian optimization is improved.

We have not discussed the theoretical aspects of Bayesian optimization (e.g., see \cite{Sri12}).  In the supplementary material, we provide some mathematical analysis in this direction, including the calculation of the fundamental {\em information gain}, and discuss difficulties in attaining regret bounds for G-Add-GP-UCB.

An interesting direction for future research is to develop variants of message passing that are targeted to continuous settings.  Our techniques can readily be applied by discretizing each continuous variable, but it would potentially be more effective (albeit non-trivial) to combine the message passing idea with a continuous global optimization procedure such as DiRect~\cite{Jones1993}.

\section{Acknowledgements}
This project has received funding from the European Research Council (ERC) under the European Union's Horizon 2020 research and innovation programme (grant agreement n$^\circ 725594$ - time-data). This work was also supported by the Swiss National Science Foundation (SNSF) under  grant number 407540\_167319.

\medskip
\bibliographystyle{unsrt}
\bibliography{ref.bib,refs_other.bib}

\begin{thebibliography}{10}

\bibitem{NIPS2012_4522}
Jasper Snoek, Hugo Larochelle, and Ryan~P Adams.
\newblock Practical {B}ayesian optimization of machine learning algorithms.
\newblock In {\em Advances in Neural Information Processing Systems}, pages
  2951--2959. 2012.

\bibitem{bergstra:hal-00642998}
James Bergstra, R.~Bardenet, Yoshua Bengio, and Bal{\'a}zs K{\'e}gl.
\newblock {Algorithms for hyper-parameter optimization}.
\newblock In {\em {Conference on Neural Information Processing Systems
  (NIPS)}}, volume~24, Granada, Spain, December 2011.

\bibitem{Mahendran:2012}
Nimalan Mahendran, Ziyu Wang, Firas Hamze, and Nando de~Freitas.
\newblock Adaptive {MCMC} with {Bayesian} optimization.
\newblock {\em Journal of Machine Learning Research - Proceedings Track for
  Artificial Intelligence and Statistics (AISTATS)}, 22:751--760, 2012.

\bibitem{07ijcai-gait}
Daniel Lizotte, Tao Wang, Michael Bowling, and Dale Schuurmans.
\newblock Automatic gait optimization with {G}aussian process regression.
\newblock In {\em International Joint Conference on Artificial Intelligence
  (IJCAI)}, pages 944--949, 2007.

\bibitem{Martinez-Cantin-RSS-07}
R.~Martinez-Cantin, N.~de~Freitas, A.~Doucet, and J.~Castellanos.
\newblock Active policy learning for robot planning and exploration under
  uncertainty.
\newblock In {\em Proceedings of Robotics: Science and Systems}, Atlanta, GA,
  USA, June 2007.

\bibitem{ComputerVision}
D.~Yamins Bergstra, J. and D.~D. Cox.
\newblock {Making a Science of Model Search: hyper-parameter optimization in
  hundreds of dimensions for vision architectures.}
\newblock In {\em {International Conference on Machine Learning (ICML),
  Atlanta, Gerorgia, 2013)}}, 2013.

\bibitem{Biology}
David C. James Neil D.~Lawrence Javier~González, Joseph~Longworth.
\newblock {B}ayesian optimization for synthetic gene design.
\newblock {\em arXiv:1505.01627}, 2015.

\bibitem{Talk}
Nando. de~Freitas.
\newblock Talk on current challenges and open problems in {B}ayesian
  optimization.
\newblock 2015.

\bibitem{Sha16}
Bobak Shahriari, Kevin Swersky, Ziyu Wang, Ryan~P Adams, and Nando de~Freitas.
\newblock Taking the human out of the loop: A review of {B}ayesian
  optimization.
\newblock {\em Proc. IEEE}, 104(1):148--175, 2016.

\bibitem{citeulike:2561921}
Peter Auer.
\newblock {Using confidence bounds for exploitation-exploration trade-offs}.
\newblock {\em J. Mach. Learn. Res.}, 3:397--422, March 2003.

\bibitem{Jones1998}
Donald~R. Jones, Matthias Schonlau, and William~J. Welch.
\newblock Efficient global optimization of expensive black-box functions.
\newblock {\em Journal of Global Optimization}, 13(4):455--492, 1998.

\bibitem{Mockus1994}
Jonas Mockus.
\newblock Application of {B}ayesian approach to numerical methods of global and
  stochastic optimization.
\newblock {\em Journal of Global Optimization}, 4(4):347--365, 1994.

\bibitem{DBLP:journals/corr/abs-1012-2599}
Eric Brochu, Vlad~M. Cora, and Nando de~Freitas.
\newblock A tutorial on {B}ayesian optimization of expensive cost functions,
  with application to active user modeling and hierarchical reinforcement
  learning.
\newblock {\em CoRR}, abs/1012.2599, 2010.

\bibitem{Hen12}
Philipp Hennig and Christian~J Schuler.
\newblock Entropy search for information-efficient global optimization.
\newblock {\em J. Mach. Learn. Research}, 13(1):1809--1837, 2012.

\bibitem{Her14}
Jos{\'e}~Miguel Hern{\'a}ndez-Lobato, Matthew~W Hoffman, and Zoubin Ghahramani.
\newblock Predictive entropy search for efficient global optimization of
  black-box functions.
\newblock In {\em Adv. Neur. Inf. Proc. Sys. (NIPS)}, pages 918--926, 2014.

\bibitem{Chen2012BO}
Andreas~Krause Bo~Chen, Rui~Castro.
\newblock Joint optimization and variable selection of high-dimensional
  gaussian processes.
\newblock {\em arXiv:1206.6396}, 2012.

\bibitem{wang2013bayesian}
Ziyu Wang, Masrour Zoghi, Frank Hutter, David Matheson, N~Freitas, et~al.
\newblock {B}ayesian optimization in high dimensions via random embeddings.
\newblock International Joint Conferences on Artificial Intelligence (AAAI),
  2013.

\bibitem{NIPS2013_5152}
Josip Djolonga, Andreas Krause, and Volkan Cevher.
\newblock High-dimensional {G}aussian process bandits.
\newblock In {\em Conference on Neural Information Processing Systems (NIPS)},
  pages 1025--1033. 2013.

\bibitem{conf/icml/KandasamySP15}
Kirthevasan Kandasamy, Jeff~G. Schneider, and Barnabas Poczos.
\newblock High dimensional {B}ayesian optimisation and bandits via additive
  models.
\newblock In {\em International Conference on Machine Learning}, volume~37,
  pages 295--304, 2015.

\bibitem{pmlr-v51-li16e}
Chun-Liang Li, Kirthevasan Kandasamy, Barnabas Poczos, and Jeff Schneider.
\newblock High dimensional {B}ayesian optimization via restricted projection
  pursuit models.
\newblock In {\em International Conference on Artificial Intelligence and
  Statistics}, Cadiz, Spain, 2016.

\bibitem{wang2017batched}
Zi~Wang, Chengtao Li, Stefanie Jegelka, and Pushmeet Kohli.
\newblock Batched high-dimensional {B}ayesian optimization via structural
  kernel learning.
\newblock {\em arXiv preprint arXiv:1703.01973}, 2017.

\bibitem{pmlr-v54-gardner17a}
Jacob Gardner, Chuan Guo, Kilian Weinberger, Roman Garnett, and Roger Grosse.
\newblock {Discovering and exploiting additive structure for {B}ayesian
  optimization}.
\newblock In {\em International Conference on Artificial Intelligence and
  Statistics}, Fort Lauderdale, FL, USA, 20--22 Apr 2017.

\bibitem{ravikumar2007spam}
Pradeep Ravikumar, Han Liu, John Lafferty, and Larry Wasserman.
\newblock Spam: Sparse additive models.
\newblock In {\em Conf. Neur. Inf. Proc. Sys. (NIPS)}. Curran Associates Inc.,
  2007.

\bibitem{tyagi17algorithms}
Hemant Tyagi, Anastasios Kyrillidis, Bernd G\"artner, and Andreas Krause.
\newblock Algorithms for learning sparse additive models with interactions in
  high dimensions.
\newblock {\em Information and Inference}, 00:1--67, August 2017.

\bibitem{hoang2018decentralized}
Trong~Nghia Hoang, Quang~Minh Hoang, Ruofei Ouyang, and Kian~Hsiang Low.
\newblock Decentralized high-dimensional bayesian optimization with factor
  graphs.
\newblock http://arxiv.org/abs/1711.07033v3, 2018.

\bibitem{icml2010_SrinivasKKS10}
Niranjan Srinivas, Andreas Krause, Matthias Seeger, and Sham~M. Kakade.
\newblock Gaussian process optimization in the bandit setting: No regret and
  experimental design.
\newblock In {\em International Conference on Machine Learning (ICML)}, pages
  1015--1022, 2010.

\bibitem{Wainwright2015}
Martin~J. Wainwright.
\newblock {\em Graphical Models and Message-Passing Algorithms: Some
  Introductory Lectures}, pages 51--108.
\newblock Springer International Publishing, Cham, 2015.

\bibitem{arnborg1987complexity}
Stefan Arnborg, Derek~G Corneil, and Andrzej Proskurowski.
\newblock Complexity of finding embeddings in ak-tree.
\newblock {\em SIAM Journal on Algebraic Discrete Methods}, 8(2):277--284,
  1987.

\bibitem{cano1995heuristic}
Andr{\'e}s Cano and Seraf{\'\i}n Moral.
\newblock Heuristic algorithms for the triangulation of graphs.
\newblock {\em Advances in Intelligent Computing (IPMU)}, pages 98--107, 1995.

\bibitem{ROBERTS1994207}
G.O. Roberts and A.F.M. Smith.
\newblock Simple conditions for the convergence of the {G}ibbs sampler and
  {M}etropolis-{H}astings algorithms.
\newblock {\em Stochastic Processes and their Applications}, 49(2):207 -- 216,
  1994.

\bibitem{Viola01rapidobject}
Paul Viola and Michael Jones.
\newblock Rapid object detection using a boosted cascade of simple features,
  2001.

\bibitem{Bradski:2008:LOE:1461412}
Dr. Gary~Rost Bradski and Adrian Kaehler.
\newblock {\em Learning {O}pen{CV}, 1st Edition}.
\newblock O'Reilly Media, Inc., first edition, 2008.

\bibitem{Sri12}
N.~Srinivas, A.~Krause, S.M. Kakade, and M.~Seeger.
\newblock Information-theoretic regret bounds for {G}aussian process
  optimization in the bandit setting.
\newblock {\em IEEE Trans. Inf. Theory}, 58(5):3250--3265, May 2012.

\bibitem{Jones1993}
D.~R. Jones, C.~D. Perttunen, and B.~E. Stuckman.
\newblock Lipschitzian optimization without the {L}ipschitz constant.
\newblock {\em Journal of Optimization Theory and Applications},
  79(1):157--181, 1993.

\bibitem{Ras06}
Carl~Edward Rasmussen.
\newblock Gaussian processes for machine learning.
\newblock MIT Press, 2006.

\bibitem{Kandasamy-arxiv}
Kirthevasan Kandasamy, Jeff~G. Schneider, and Barnabás Póczos.
\newblock High dimensional {B}ayesian optimisation and bandits via additive
  models.
\newblock {\em arXiv:1503.01673 [v3]}, 2017.

\bibitem{4170}
MW. Seeger, SM. Kakade, and DP. Foster.
\newblock Information consistency of nonparametric gaussian process methods.
\newblock {\em IEEE Transactions on Information Theory}, 54(5):2376--2382, May
  2008.

\end{thebibliography}

\appendix

\onecolumn

\newpage

{\centering
{\huge \bf Supplementary Material}

{\Large \bf High-Dimensional Bayesian Optimization via Additive Models \\ with Overlapping Groups (AISTATS 2018) \par } 

 {\large Paul Rolland, Jonathan Scarlett, Ilija Bogunovic, and Volkan Cevher } 

}

\bigskip

All citations below are to the reference list in the main document.

\section{Derivation of posterior mean and variance}

In this section, we prove equation~\eqref{AddPosterior} for the posterior mean and variance conditioned on the observations. The derivation is similar to that of Kandasamy {\em et al.}~\cite{conf/icml/KandasamySP15}, which in turn is similar to the standard Gaussian process posterior derivation \cite{Ras06}.

Recall that instead of directly computing the posterior mean and variance on the high-dimensional function, we are considering the terms $f^{(j)}$ in the additive decomposition of $f$ separately. We are thus interested in the distributions of $f_*^{(j)} = f^{(j)}(x_*^{(j)})$, $j=1,...,M$ conditioned on the noisy samples $\yv = {y_1,...,y_n}$ at points $\xv = {x_1,...,x_n}$, for some query points $x_*^{(j)}$. We claim that the joint distribution of $f_*^{(j)}$ and $\yv$ can be written as
\begin{equation}
\begin{pmatrix}
f_*^{(j)} \\ \yv
\end{pmatrix}
\sim \mathcal{N} \left(\textbf{0},
\begin{pmatrix}
\kappa ^{(j)}(x_*^{(j)}, x_*^{(j)}) & \kappa ^{(j)}(x_*^{(j)}, \xv^{(j)}) \\
\kappa ^{(j)}(\xv^{(j)}, x_*^{(j)}) & \kappa (\xv,\xv) + \eta ^2I_n
\end{pmatrix}
\right) \label{eq:posterior_joint}
\end{equation}
To see this, we recall that distinct functions in the additive decomposition~\eqref{fdecomp} are independent given $\xv$. Hence, for any observation $y_p = f(x_p) + \epsilon$ and $j=1,...,M$, we have
\begin{align*}
\mathrm{Cov}(f_*^{(j)}, y_p) &= \mathrm{Cov}\left(f_*^{(j)}, \sum_{i=1}^M f^{(i)}(x_p^{(i)}) + \epsilon\right) \\
&= \mathrm{Cov}\left(f_*^{(j)}, f^{(j)}(x_p^{(j)})\right) \\
&= \kappa^{(j)}(x_*^{(j)}, x_p^{(j)}),
\end{align*}
which establishes \eqref{eq:posterior_joint}.

With \eqref{eq:posterior_joint} in place, we can use a standard conditional Gaussian formula (as used in standard GP posterior derivations \cite{Ras06}, as well as the non-overlapping setting of \cite{conf/icml/KandasamySP15}) to derive the posterior mean and variance. Specifically, defining the matrix $\Delta = \kappa(\xv,\xv) + \eta^2I_n \in \mathbb{R}^{n\times n}$, we have for past query points $\xv$ and next query point $x_*^{(j)}$ that
\begin{equation}
\begin{split}
(f_*^{(j)}|\yv) \sim \mathcal{N}&\left(\kappa^{(j)}(x_*^{(j)}, \xv^{(j)})\Delta^{-1}\yv,  \right.\\
&\left. \kappa^{(j)}(x_*^{(j)}, x_*^{(j)}) - \kappa^{(j)}(x_*^{(j)},\xv^{(j)})\Delta^{-1}\kappa^{(j)}(\xv^{(j)}, x_*^{(j)}) \right)
\end{split}
\label{sumPosterior}
\end{equation}
under the notation in \eqref{AddPosterior}.  This concludes the derivation.


\section{Mathematical analysis and theoretical challenges}

\subsection{Discussion on existing theory}

{\bf Guarantees of GP-UCB.} A notable early work providing theoretical guarantees on Bayesian optimization (without the high-dimensional aspects) is that of Srinivas {\em et al.}~\cite{Sri12}, who considered the {\em cumulative regret} $R_T = \sum_{t=1}^T \big( f(x_{\mathrm{opt}}) - f(x_t)\big)$ with $x_{\mathrm{opt}} = \argmax_{x\in \Xc} f(x)$.  In the case of a finite domain $\Xc$, it was shown that the GP-UCB algorithm with exploration parameter $\beta_t = 2\log\big( \frac{|\Xc| t^2\pi^2}{6\delta} \big)$ achieves
\begin{equation}
R_T \le \sqrt{\frac{8}{\log(1+\sigma^{-2})} T\beta_T\gamma_T} \label{eq:srinivas}
\end{equation}
with probability at least $1-\delta$.  Here the kernel-dependent quantity $\gamma_T$ is known as the {\em information gain}, and is defined as the maximum of a mutual information quantity:
\begin{equation}
\gamma_T = \max_{A \,:\, |A|=T} I(y_A;f_A) \label{eq:gamma}
\end{equation}
for $A = \{x_1,...,x_T\}$ with corresponding function values $f_A$ and observations $y_A$.  Analogous results were presented for the continuous setting in \cite{Sri12} under mild technical assumptions, and explicit bounds on $\gamma_T$ were provided for the squared exponential and Mat\'ern kernels.

{\bf Extension to non-overlapping additive models.} Kandasamy {\em et al.}~\cite{conf/icml/KandasamySP15} attempted to upper bound the cumulative regret of their algorithm Add-GP-UCB in the high-dimensional setting with additive models. In particular, they sought a bound whose complexity is only exponential in the maximal dimension $d$ of the low-dimensional kernels, instead of the full dimension $D$ as in Srinivas {\em et al.}~\cite{Sri12}. However, they subsequently stated in an updated version of their paper that the proof contains an error~\cite{Kandasamy-arxiv}.  In our understanding, the error is due to the fact that the approximated standard deviation $\sum_{i=1}^M \sigma_{t-1}^{(i)}$ is different from the true standard deviation $\sigma_{t-1}$, and that the ratio $\frac{\sum_{i=1}^M \sigma_{t-1}^{(i)}(x)}{\sigma_{t-1}(x)}$ cannot be upper bounded for all $x$ (see below).

In the parallel independent work of Hoang {\em et al.} \cite{hoang2018decentralized}, it was shown that a sufficient condition that leads to sub-linear regret bounds in high-dimensional additive models is as follows \cite[Assumption 4]{hoang2018decentralized}: The posterior variance $\sigma_t^{(i)}(x^{(i)})$ of each component $i$ given the observations can be upper bounded by a constant times $\widehat{\sigma}_t^{(i)}(x^{(i)})$, defined to be the posterior variance as if the corresponding function $f^{(i)}$ had been sampled directly instead.  However, it remains an open problem to determine specific models and kernels for which this assumption is true.

{\bf Outline of this appendix.} We further discuss the relation between the true and approximate posterior standard deviations in Section \ref{sec:var_relation}, and then provide a novel bound on the information gain for our setting in Section \ref{sec:gamma_bound}.  While the latter is only one step towards attaining a regret bound for G-Add-GP-UCB, it also quantifies the regret bound \eqref{eq:srinivas} when GP-UCB is applied to the high-dimensional setting.  Unfortunately, GP-UCB is not computationally feasible in high dimensions, so establishing a similar regret bound G-Add-GP-UCB remains an important direction for future research.

\subsection{Relation between true posterior variance and its approximation} \label{sec:var_relation}

Our algorithm G-Add-GP-UCB is based on an acquisition function that can be computed efficiently in high dimensions. This property comes from the fact that it that can be decomposed into the sum of low-dimensional components (see \eqref{eq:acq_func}). Each term in the sum consists of a mean and standard deviation corresponding to a low-dimensional function.

We observe that $\tilde{\mu}_{t-1}(x) = \sum_{j=1}^M \mu_{t-1}^{(j)}(x^{(j)}) = \mu_{t-1}(x)$. Therefore, this way of splitting the posterior mean into several lower dimensional components does not involve any approximation. However, $\tilde{\sigma}_{t-1}(x) = \sum_{j=1}^M \sigma_{t-1}^{(j)}(x^{(j)}) \neq \sigma_{t-1}(x)$ in general; this can be viewed as being due to the non-linearity of the quadratic term $\kappa(x_*,\xv)\Delta^{-1}\kappa(\xv, x_*)$ in the posterior variance \eqref{AddPosterior}.

Our analysis below reveals that
\begin{equation}
\sum_{j=1}^M \sigma_{t-1}^{(j)}(x^{(j)}) \geq \sigma_{t-1}(x). \label{VarianceApproximationBound}
\end{equation}
Thus, this splitting of the posterior standard deviation into low-dimensional components generally over-estimates the true variance. An example where the inequality is strict is as follows: In the case of zero noise, the true posterior standard deviation at a point $x_{\mathrm{evaluated}}$ that has already been evaluated is $\sigma_{t-1}(x_{\mathrm{evaluated}}) = 0$. However in general, $\sum_{j=1}^M \sigma_{t-1}^{(j)}(x_{\mathrm{evaluated}}) > 0$. Therefore, the ratio $\frac{\sum_{j=1}^M \sigma_{t-1}^{(j)}(x)}{\sigma_{t-1}(x)}$ can sometimes diverge, and it is thus not possible to upper bound it for all $x$.

{\bf Derivation of the upper bound \eqref{VarianceApproximationBound}.}  The true posterior variance based on observations at the points $\xv=(x_1,...,x_t)$ is given by
\begin{equation}
\sigma_t(x)^2 = \kappa(x,x) - \kappa(x,\xv)\Delta^{-1}\kappa(\xv,x),
\end{equation}
where $\kappa$ is the full dimensional kernel $\kappa(x,x') = \sum_{i=1}^M \kappa^{(i)}(x^{(i)}, x'^{(i)})$, $k(x,\xv)$ and $k(\xv,x)$ are the corresponding vectors of kernel values, and $\Delta$ is a matrix such that $\Delta_{ij} = \kappa(x_i, x_j)$ for $i,j = 1,...,t$. The approximated posterior variance based on the same points is as follows:
\begin{align}
\sum_{i=1}^M \sigma_t(x^{(i)})^2 &= \sum_{i=1}^M \kappa^{(i)}(x^{(i)}, x^{(i)}) - \kappa^{(i)}(x^{(i)},\xv^{(i)})\Delta^{-1}\kappa^{(i)}(\xv^{(i)},x^{(i)}) \\
&= \kappa(x,x) - \sum_{i=1}^M \kappa^{(i)}(x^{(i)},\xv^{(i)})\Delta^{-1}\kappa^{(i)}(\xv^{(i)},x^{(i)})
\end{align}
under the notation following \eqref{AddPosterior}.

By definition, the matrix $\Delta$ is symmetric and positive definite, and hence so is the matrix $\Delta^{-1}$. We can thus define a norm induced by this matrix on the space $\mathbb{R}^t$; for $\vec{k} \in \mathbb{R}^t$, we have
\begin{equation}
\|\vec{k}\|_{\Delta^{-1}}^2 = \vec{k}^T\Delta^{-1} \vec{k}.
\end{equation}

The fact that $\Delta^{-1}$ is symmetric positive definite implies that this has all the properties of a norm. For any $x\in \mathbb{R}^D$, we define the $t$-dimensional vector $\vec{k}^{(i)}(x)$ as $\vec{k}^{(i)}(x)_j = \kappa^{(i)}(x^{(i)}, x_j^{(i)})$. We also recall that $\kappa(x,x) = 1$ for all $x\in \mathbb{R}^D$. Using this notation, we can rewrite the expressions for the true and approximated posterior variances as
\begin{equation}
\sigma_t(x)^2 = 1 - \Big\|\sum_{i=1}^M\vec{k}^{(i)}(x)\Big\|^2_{\Delta^{-1}}
\end{equation}
and
\begin{equation}
\sum_{i=1}^M \sigma_t(x^{(i)})^2 = 1 - \sum_{i=1}^M\|\vec{k}^{(i)}(x)\|^2_{\Delta^{-1}}.
\end{equation}

By the triangle inequality, we have
\begin{align*}
\sigma_t(x)^2 &= 1 - \Big\|\sum_{i=1}^M\vec{k}^{(i)}(x)\Big\|^2_{\Delta^{-1}} \\
&\leq 1 - \bigg(\sum_{i=1}^M\|\vec{k}^{(i)}(x)\|_{\Delta^{-1}}\bigg)^2 \\
&\leq 1 - \sum_{i=1}^M\|\vec{k}^{(i)}(x)\|^2_{\Delta^{-1}} \\
&=  \sum_{i=1}^M \sigma_t(x^{(i)})^2
\end{align*}
As $\sigma_t(x^{(i)}) \geq 0$ $\forall x\in \mathbb{R}^D$, we have that $\sum_{i=1} \sigma_t(x^{(i)})^2 \leq \left(\sum_{i=1} \sigma_t(x^{(i)})\right)^2$, which implies
\begin{equation}
\sigma_t(x) \leq \sum_{i=1}^M \sigma_t(x^{(i)})
\end{equation}
as desired.

%

\begin{figure}[h!]
\centering
\begin{subfigure}{.45\textwidth}
\centering
\includegraphics[width=1.\linewidth]{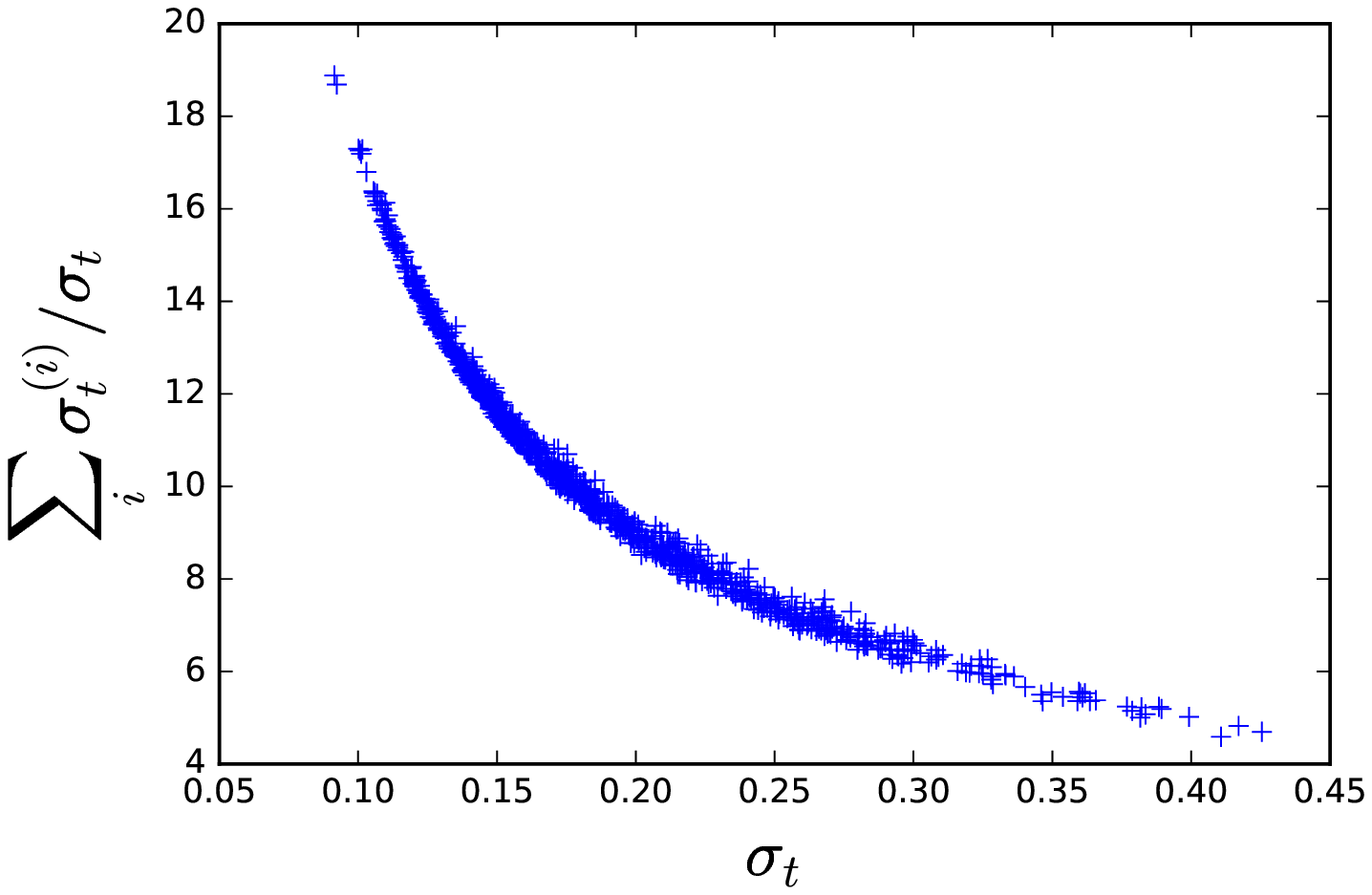}
\end{subfigure}
\begin{subfigure}{.45\textwidth}
\centering
\includegraphics[width=1.\linewidth]{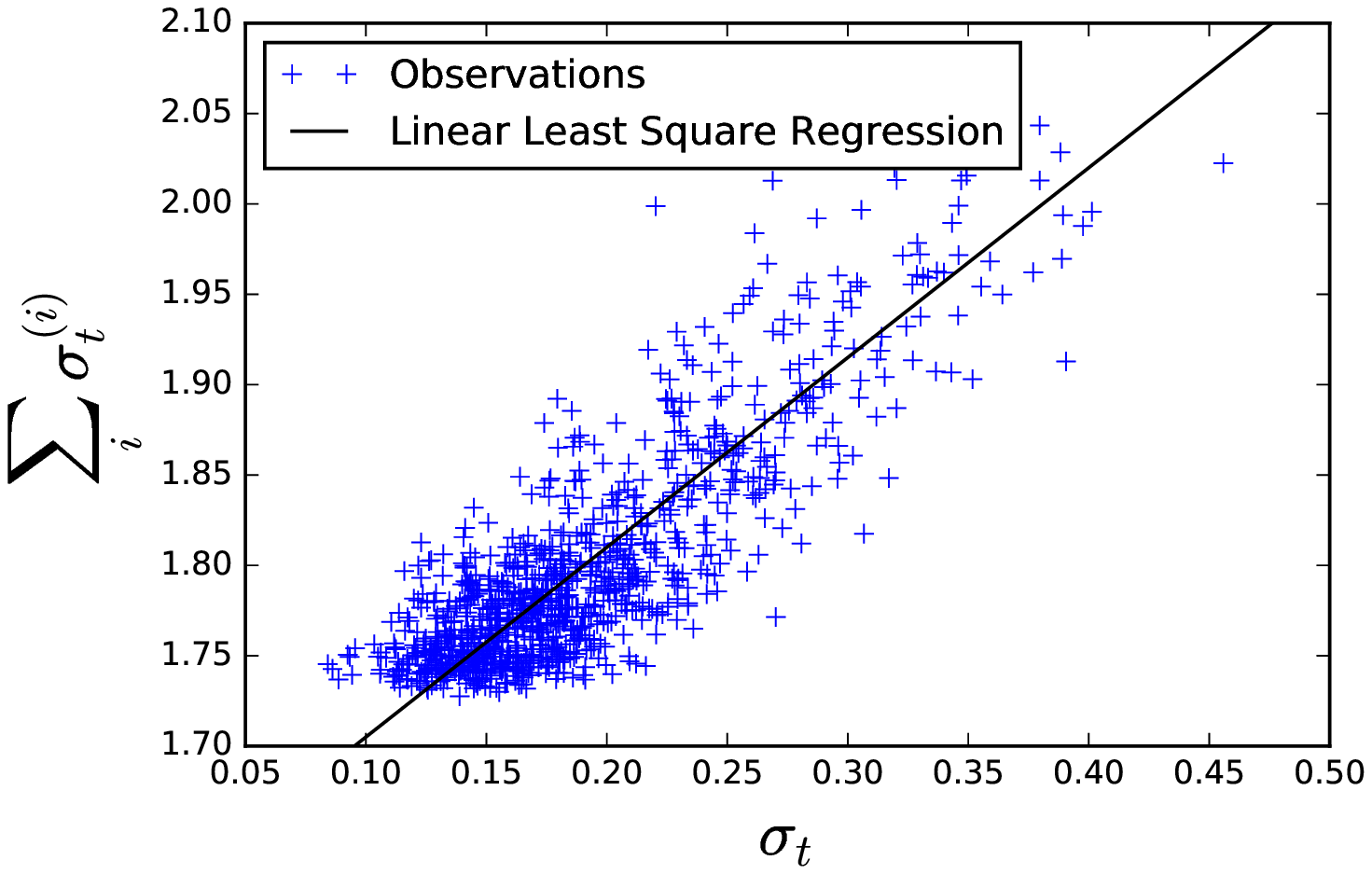}
\end{subfigure}
\caption{Evaluation of the posterior standard deviation from the full dimensional kernel (true posterior variance), vs.~separating the kernel into lower dimensional additive components (approximated posterior variance). Left: Ratio between the two computed posterior standard variations as a function of the true posterior standard variation. Right: Approximated posterior standard deviation versus true posterior standard deviation.}
\label{VarianceAnalysis}
\end{figure}

{\bf Numerical evaluation.}  In order to observe the difference between the true posterior variance $\sigma_t^2$ and the approximated variance $\left(\sum_{j=1}^M \sigma_{t-1}^{(j)}\right)^2$, we generate a $10$ dimensional synthetic function via Gaussian processes with the star dependency graph shown in Figure~\ref{SyntheticDepGraphs}. We first evaluate this function at $200$ randomly-selected points. Based on these observations, we then evaluate the true and approximated posterior variances at $1000$ randomly selected points. We then compare the results obtained with these two different methods (Figure~\ref{VarianceAnalysis}).

From the figure on the right, we can observe that the obtained values can be significantly different. However, there is a clear correlation between these two quantities, in the sense that points with higher true variance tend to also have a higher approximated variance. The figure on the left shows that the ratio between approximated and true variance increases as the true variance becomes smaller. This approximation error strongly depends on the decomposition, and in particular on $M$. The higher the value of $M$, the higher the approximation error.


\subsection{Bounding the information gain with overlapping groups} \label{sec:gamma_bound}

As outlined above, the information gain $\gamma_T$ in \eqref{eq:gamma} plays a crucial role in the regret bounds for Bayesian optimization.  While we do not claim any regret bounds for G-Add-GP-UCB, bounding $\gamma_T$ may provide an initial step towards this, and also allows us to understand the performance of GP-UCB with our structured kernels ({\em cf.}, \eqref{eq:srinivas}).  We provide such a bound for our setting, focusing on the squared exponential kernel, and ultimately showing that $\gamma_T = O(Dd^d(\log{T})^{d+1})$ analogously to \cite{conf/icml/KandasamySP15}.  Here $d$ is the highest dimension of any low-dimensional function in the additive decomposition.

We follow the high-level steps of \cite{conf/icml/KandasamySP15} with suitable modifications for our setting.  It was shown in \cite{icml2010_SrinivasKKS10} that under some mild assumptions on the target function $f$, the maximal information gain can be bounded as
\begin{align}
\gamma_T \leq \inf\limits_{\tau} &\left( \frac{1/2}{1-e^{-1}}\max_{r\in\{1,...,T\}}(T_*\log(rn_T/\eta^2) +  C_9\eta^2(1-\frac{r}{T})(T^{r+1}B_{\kappa}(T_*)+1)\log{T}) + O(T^{1-\tau/D})\right),
\label{bigBound}
\end{align}
for any $T_*\in\{1,...,\min(T,n_T)\}$, where $C_9 = 4D+2$, $n_T = C_9T^{\tau}\log{T}$, and $B_{\kappa}(T_*) = \sum_{s > T_*} \lambda_s$. Here $\{\lambda_n\}_{n\in \mathbb{N}}$ are the eigenvalues of $\kappa$ with respect to the uniform distribution.

In order to bound $\gamma_T$, it therefore suffices to bound $B_{\kappa}(T_*)$, i.e., to bound the sum of the eigenvalues of $\kappa$ at the tail. Unlike the setting of \cite{conf/icml/KandasamySP15}, the eigenfunctions corresponding to different kernels $\kappa^{(i)}$ and $\kappa^{(j)}$ are not necessarily orthogonal, since overlaps between kernel variables are possible. To circumvent this difficulty, we can make use of Weyl's inequality.
\begin{lem}
{\em (Weyl's inequality)}
Let $H, P \in \mathbb{R}^{n\times n}$ be two Hermitian matrices, and define $M=H+P$. Let $\mu_i, \nu_i, \rho_i$, $i=1,...,n$ be the eigenvalues of $M$, $H$ and $P$ respectively such that $\mu_1 \geq ... \geq \mu_n$, $\nu_1 \geq ... \geq \nu_n$ and $\rho_1 \geq ... \geq \rho_n$. Then for all $i\geq r+s-1$, we have
\begin{equation}
\mu_i \leq \nu_r + \rho_s
\end{equation}
\end{lem}

This result immediately generalizes to a sum with an arbitrary number of matrices. In particular, we will use the following corollary.
\begin{corollary}
Let $K_i \in \mathbb{R}^{n\times n}$, $i=1,...,M$ be Hermitian matrices, and define $K = \sum_{i=1}^M K_i$. Let $\{\lambda^{(i)}_j\}_{j=1,...,n}$, be the eigenvalues of $K_i$ such that $\lambda^{(i)}_1 \geq ... \geq \lambda^{(i)}_n$ $\forall i=1,...,M$, and let $\{\lambda_i\}_{i=1,...,n}$ be the eigenvalues of $K$ such that $\lambda_1 \geq ... \geq \lambda_n$. Then for all $k\in \mathbb{N}$ such that $kM+1 \leq n$, we have
\begin{equation}
\lambda_{kM+1} \leq \sum_{i=1}^M \lambda^{(i)}_{k+1}.
\end{equation}
\end{corollary}

Let $\{\lambda_s\}_{s\in \mathbb{N}}$, $\lambda_1 \geq \lambda_2\geq ...$ denote the eigenvalues of $\kappa$, and for all $j=1,...,M$, let $\{\lambda_s^{(j)}\}_{s\in \mathbb{N}}$, $\lambda_1^{(j)} \geq \lambda_2^{(j)}\geq ...$ denote the eigenvalues of $\kappa^{(j)}$. It was shown by Seeger {\em et al.}~\cite{4170} that the eigenvalues $\lambda_s^{(j)}$ for the square exponential kernel $\kappa^{(j)}$ satisfy $\lambda_s^{(j)} \leq c^dB^{s^{1/d}}$, $B < 1$, where each $\kappa^{(j)}$ is a $d_j$-dimensional kernel, and $d_j \leq d$. Defining $T_+ = \left \lfloor\frac{T_*}{M}\right\rfloor$ we have the following:
\begin{align}
B_{\kappa}(T_*) &= \sum_{s > T_*} \lambda_s \\
&\leq \sum_{k > T_+} \sum_{i=1}^{M} \lambda_{(k-1)M + l} \\
&\leq \sum_{k > T_+} \sum_{i=1}^{M} \sum_{j=1}^M \lambda_k^{(j)} \\
&\leq M^2c^d\sum_{k > T_+} B^{k^{1/d}},
\end{align}
where the second line uses the fact that the eigenvalues are increasingly ordered, the third line follows from Weyl's inequality, and the final line follows from the bound on the tail eigenvalues given in \cite{4170}.

The rest of the proof follows via a similar analysis to~\cite{conf/icml/KandasamySP15}. One difference is that we get an extra $M$ term in our bound for $B_{\kappa}$ compared to the setting of \cite{conf/icml/KandasamySP15}. However, this does not affect the bound for $\gamma_T$, since the leading term on the right hand side of~\eqref{bigBound} is $T_*\log(rn_T/\eta^2)$ which does not involve $B_{\kappa}$. We thus obtain the same bound for $\gamma_T$ as in~\cite{conf/icml/KandasamySP15}, namely,
\begin{equation}
\gamma_T = O(Dd^d(\log{T})^{d+1}).
\end{equation}
Note that this bound only has linear dependence on the dimension $D$, while being exponential in the maximal group size $d$.

\section{Astrophysical data experiment}

In this appendix, we consider an additional experiment on real-world data that aims at estimate a set of $9$ cosmological parameters (e.g., Hubble's constant, proportion of helium in the universe, etc) in order to best match reality. These constants are involved in the theoretical model of physics, but are estimated experimentally. To do so, programs model the dynamics of the universe given these parameters, and compare the results of the simulations with the observed data.

For each set of parameters, we can compute the likelihood that the chosen parameters match the reality. The aim is thus to find the set of parameters that maximize this likelihood, or equivalently that minimize the negative log-likelihood. We use the LRG DR7 Likelihood Software released by NASA\footnote{http://lambda.gsfc.nasa.gov/toolbox/lrgdr/} in order to compute likelihoods given these cosmological parameters based on experimental data released by the Sloan Digital Sky Survey.

We note that this data was used by both Kandasamy {\em et al.}~\cite{conf/icml/KandasamySP15} and Gardner {\em et al.}~\cite{pmlr-v54-gardner17a} for testing high-dimensional BO algorithms, but it was used in somewhat different ways.  We adopt the approach of \cite{pmlr-v54-gardner17a}, and we avoid adding additional ``dummy dimensions'' as in \cite{conf/icml/KandasamySP15}.

The software provides a set of parameters which achieves a negative log-likelihood of $23.7$. We thus apply the two Bayesian algorithms ``Overlap'' and ``No Overlap'' in a range of $75\%-125\%$ of this default set of parameters. Unlike the previous real world experiment, we do not set a fixed dependency graph for the ``Overlap'' model and learn it throughout the algorithm using Gibbs sampling ({\em cf.}, Section \ref{sec:learn_graph}).  Similarly, for the ``No Overlap'' model, we use the Gibbs sampling approach of \cite{wang2017batched}, placing no ``hard constraints'' (i.e., choices of $M$ and $d$ in \cite{conf/icml/KandasamySP15}).

\begin{figure}[h!]
\centering
\includegraphics[width=0.8\linewidth]{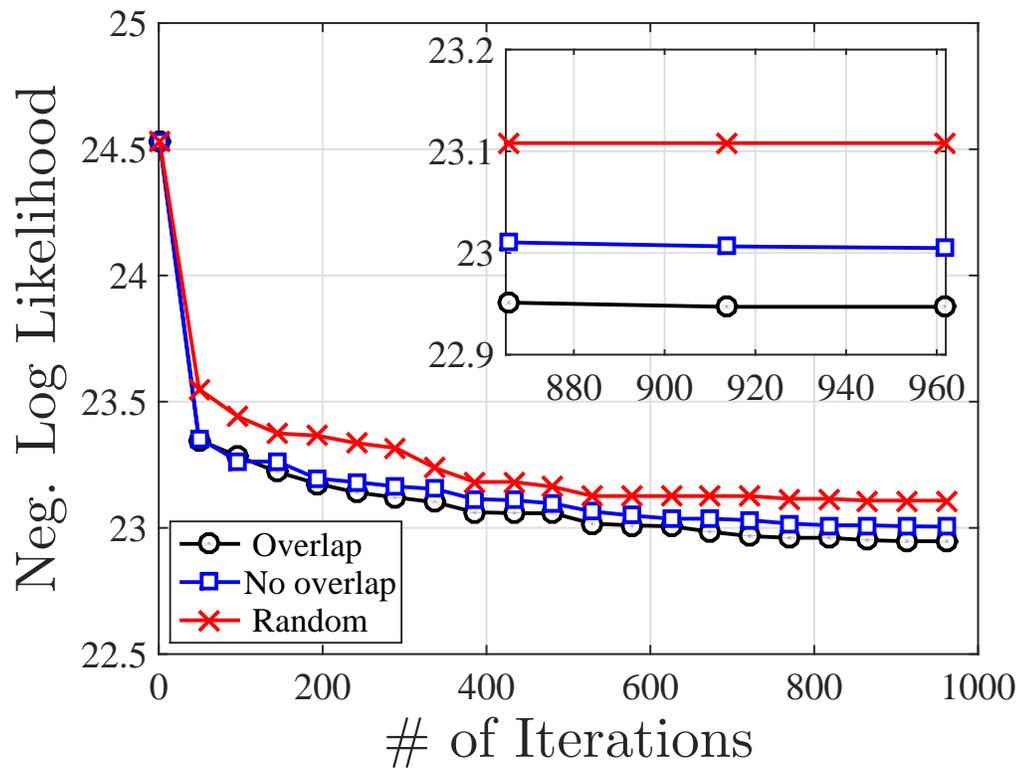}
\caption{Results on the astrophysical experiment. The lower the vertical axis value, the more likely it is that the chosen constants match the observed data.}
\label{fig:SDSS_Results}
\end{figure}

The remaining parameters for the Bayesian optimization algorithms are the same as for the first real-world experiment (Table~\ref{tab:ParamRealBO}), except that we run the simulation for 1000 iterations.  The results are shown in Figure~\ref{fig:SDSS_Results}.  We observe that both algorithms achieve a higher likelihood than with the default parameters, and that the ``Overlap'' algorithm achieves a higher likelihood than the ``No Overlap'' one across the entire time horizon.  In particular, by the final iteration, the gap to ``random'' has increased from approximately $0.09$ to $0.15$, i.e., an increase of over $60\%$.

\end{document}